\documentclass{article}

 \usepackage[main, final]{neurips_2026}

\usepackage[utf8]{inputenc} 
\usepackage[T1]{fontenc}    
\usepackage{hyperref}       
\usepackage{url}            
\usepackage{booktabs}       
\usepackage{amsfonts}       
\usepackage{nicefrac}       
\usepackage{microtype}      
\usepackage{xcolor}         
\usepackage{graphicx}
\usepackage{amsmath}
\usepackage{multirow}
\usepackage{wrapfig}
\usepackage{subcaption}
\usepackage{placeins}

\usepackage[most]{tcolorbox}
\tcbset{
  custombox/.style={
    width=\linewidth,
    top=3pt,
    bottom=3pt,
    left=3pt,
    right=3pt,
    colback=gray!5!white,
    colframe=darkgray,
    colbacktitle=darkgray,
    enhanced,
    center,
    attach boxed title to top left={yshift=-0.12in,xshift=0.2in},
    boxed title style={
      boxrule=0pt,
      colframe=white,
      rounded corners,
      interior style={color=black}
    },
  }
}
\newtcolorbox{CustomBox}[2][]{custombox,title=#2,#1}
\definecolor{darkgray}{rgb}{0.25,0.25,0.25}

\title{Rethinking Visual Reasoning in \\ Text-to-Image Reward Modeling}

\author{%
    Shiye Su\thanks{Correspondence to \texttt{shiye@stanford.edu}. Code: \texttt{https://github.com/ssu53/crosspairs-rm}.} \\
    Stanford University
    \And
    Xiaohan Wang \\
    Stanford University
    \And 
    Ludwig Schmidt \\
    Stanford University
    \And
    Serena Yeung-Levy \\
    Stanford University
}

\begin{document}

\maketitle

\begin{abstract}
Reward models are a core component of modern text-to-image systems, serving as the objective for inference-time selection and RL-based post-training.
A key prerequisite is visual understanding: to judge prompt adherence and ground aesthetic preferences, a reward model must bind attributes, resolve relations, and count reliably.
We find that widely-used reward models perform poorly on targeted visual reasoning evaluations. 
Even when starting from strong vision–language backbones, preference fine-tuning erodes these capabilities, and continual-learning baselines fail to mitigate a steep tradeoff between visual reasoning and preference accuracy.
We propose a data-centric intervention that injects visual reasoning supervision into preference learning by \emph{flipping the pairing axis}.
To complement conventional human preference data (paired images under a shared prompt), we convert existing visual reasoning datasets into preference pairs that \emph{share the same image but differ in prompt}, and train on the mixture.
This yields a Pareto improvement in both visual understanding and preference prediction.
Across inference-time best-of-$n$ selection and RL post-training, the resulting reward model improves text-to-image generation on challenging compositional prompts (GenEval) by an average of 4.0\% and 5.5\% respectively, 
outperforming HPSv3 which was trained on 6$\times$ more preference pairs.
\end{abstract}
\section{Introduction}

\begin{figure}
\includegraphics[width=0.95\textwidth]{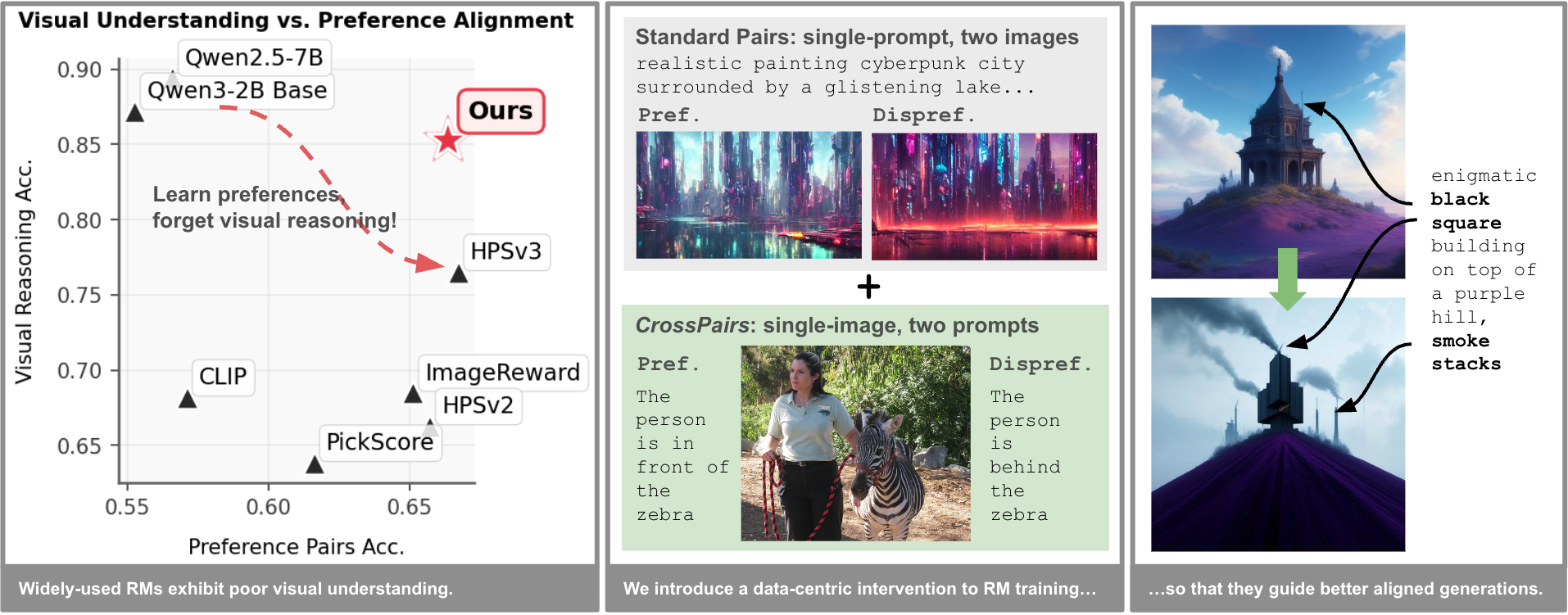}
\centering
\caption{
\textit{(Left)} Existing reward models exhibit poor visual understanding. VLM backbones possess strong visual reasoning, but this is eroded during preference fine-tuning, even with robust fine-tuning techniques. \textit{(Center)} We introduce CrossPairs -- preference pairs sharing a common image but differing in prompt -- to isolate text-image alignment signal by eliminating aesthetic confounds. Mixed with standard pairs during training, our method achieves a Pareto improvement. \textit{(Right)} The resulting reward model guides generations with superior compositional accuracy.
}
\label{fig:fig1}
\vspace{-1em}
\end{figure}

Text-to-image generative models can produce compelling visuals from natural language prompts \citep{stabilityai2024sd35,flux2024,deepmind2024imagen3}, 
yet faithfully rendering all elements — multiple objects, their attributes, spatial arrangements, and quantities — while satisfying human aesthetic preferences remains challenging \citep{rassin2023linguistic,kajic2024evaluating,huang2025t2i}. 
It is standard to utilize a reward model during post-training or inference-time selection to optimize generations towards desired behaviors \citep{ma2025inference,black2023training, wallace2024diffusion}.
A reward model must accomplish two intertwined objectives: it must assess whether the generated image aligns with the semantic guidance of the prompt, and it must capture human preferences regarding aesthetics. 
\textit{Visual understanding} is crucial to assessing alignment, and a prerequisite for grounded aesthetic judgments.
A reward model that cannot reliably parse the contents of an image, count objects, identify spatial relationships, or bind attributes to entities will struggle to judge whether the image faithfully realizes the user's intent. 

Despite the centrality of visual understanding to reward modeling, we find that popular text-to-image reward models are remarkably poor visual reasoners. 
We introduce evaluations probing compositional reasoning, positional understanding, and numeracy, revealing that models like ImageReward, PickScore, and HPS perform near chance on basic tasks. 
A natural remedy is to leverage modern vision-language models (VLMs) as backbones, given their substantially stronger compositional understanding. However,
we find that fine-tuning VLMs on preference data erodes the very capabilities that make them attractive.
Continual learning techniques — parameter-efficient fine-tuning, weight merging, and uncertainty calibration — either fail to adequately preserve visual understanding or offer unfavorable tradeoffs with preference alignment.

These findings point to a data problem.
Conventional preference datasets pair a single prompt with two images of differing quality, entangling prompt adherence with aesthetics.
Because the two images often differ simultaneously in style, artifacts, and semantic correctness, a reward model can obtain high preference accuracy by exploiting aesthetic shortcuts rather than learning robust visual parsing.
Simply scaling such data does not resolve the issue: HPSv3 is trained on 1.3M preference pairs, yet still underperforms its base model on visual understanding.
We therefore propose a data-centric intervention that isolates the alignment signal that depends on visual reasoning.
Our key insight is to \emph{flip the pairing axis}: construct preference pairs that share a \textit{common image} but \textit{differ in prompt}.
Holding the image fixed removes aesthetic confounds and forces the reward model to judge which prompt the image actually satisfies.
This formulation maps naturally onto existing visual reasoning benchmarks, which can be converted into shared-image preference pairs without additional annotation.
Finally, we mix these shared-image pairs with conventional preference data during training.

Our resulting reward model achieves a Pareto improvement, matches or exceeds the accuracy of existing reward models on held-out preference pairs while substantially outperforming them on compositional understanding tasks. We evaluate our model as a reward signal for both inference-time scaling via best-of-$n$ selection and for reinforcement learning post-training with DiffusionNFT \citep{zheng2025diffusionnft}. Our reward model strongly outperforms baselines on GenEval \citep{ghosh2023geneval}, which explicitly tests challenging prompts that require precise rendering of object and attribute relations,
by an average of 4.0\% in the best-of-$n$ setting
and an average of 5.5\% in the RL setting.
Qualitative examples from generic user prompts show that RL on our reward model yields aesthetic generations with superior prompt adherence than baselines.
\section{Preliminaries}
\label{sec:prelims}

We briefly review the role of reward models in text-to-image generation.
Let $p_\theta(x|c)$ be a pretrained generative model that produces an image $x$ conditioned on a text prompt $c$.
We consider reward models of the form $r_\phi(x,c)$ that produce a continuous scalar score of quality.
These reward models may be leveraged in \textbf{inference-time scaling} to filter, rerank, and select the best generation, e.g.
\begin{equation}
x^* = \arg\max_{\{x_1,...,x_n\}} \; r_\phi(x_i, c)
\end{equation}
where $\{x_i\}$ are candidates drawn from $p_\theta(\cdot|c)$.
More complex test-time selection schemes include optimizing $p_\theta$'s noise latents or steering its score field \citep{ma2025inference, wallace2023end, bansal2023universal}.
The reward model can also be leveraged in \textbf{reinforcement-learning post-training} to update $\theta$ towards higher reward generations: 
\begin{equation}
\max_\theta  \mathbb{E}_{c \sim \mathcal{D}, x \sim p_\theta(\cdot | c)} \left[ r_\phi(x, c) \right] - \beta \, \text{KL}\left( p_\theta(\cdot | c) \| p_{\text{ref}}(\cdot | c) \right)
\end{equation}
where $\beta$ controls the strength of regularization towards a reference model $p_\text{ref}$ (typically taken to be the pretrained generative models).
Many RL algorithms for optimizing text-to-image models exist \citep{black2023training, liu2025flow, zheng2025diffusionnft}, all relying on a reward signal $r_\phi$ to robustly guide the optimization.

Reward models are generally calibrated on datasets of \textbf{pairwise human preferences} $\mathcal{D}=\{(c,x_w,x_l)\}$ comprising a preferred ($x_w$) and dispreferred ($x_l$) generation for prompt $c$.
To train $r_\phi$, we adopt the standard Bradley-Terry formulation \citep{bradley1952rank} for human preferences, which models the probability that $x_w$ is preferred over $x_l$ as
\begin{equation}
P(x_w \succ x_l \mid c) = \sigma\left( r_\phi(x_w, c) - r_\phi(x_l, c) \right),
\label{eq:pref_diff_image}
\end{equation}
which induces a maximum likelihood objective for $r_\phi$:
\begin{equation}
\mathcal{L}(\phi) = -\mathbb{E}_{\mathcal{D}} \left[ \log \sigma\left( r_\phi(x_w, c) - r_\phi(x_l, c) \right) \right].
\end{equation}

\section{Widely-Used Reward Models are Poor Visual Reasoners}
\label{sec:diagnosis}

\begin{table*}[t]
  \caption{Accuracy of widely-used reward models on ImageRewardDB and our visual understanding evaluations. Random performance is 50\%. Bottom: HPSv3 exhibits substantially degraded visual understanding relative to the base VLM it was fine-tuned from.}
  \label{tab:acc_rms}
  \begin{center}
    \begin{footnotesize}
    \setlength{\tabcolsep}{1.5pt}
    \begin{tabular}{lccrrrrrrrr}
    \toprule
      &
      \multirow{2}{*}{\centering ImageRDB} &
      \multirow{2}{*}{\centering Visual} &
      \multicolumn{8}{c}{Visual Understanding Breakdown} \\
      \cmidrule(lr){4-11}
      & & &
      Wground & SC-Add & SC-Rep & SC-Swap & CV-Rel & WhatsUp & CV-Count & CountBch \\
    \midrule
    CLIP
      & 57.1 & 68.2 & 55.4 & 89.9 & 85.4 & 71.9 & 53.5 & 48.9 & 57.8 & 82.6 \\
    ImageReward
      & 65.1 & 68.5 & 56.8 & 96.1 & 88.4 & 83.9 & 51.8 & 52.0 & 55.6 & 63.4 \\
    PickScore
      & 61.6 & 63.8 & 54.6 & 81.6 & 80.6 & 64.9 & 51.7 & 50.6 & 52.8 & 73.8 \\
    HPSv2
      & 65.7 & 66.3 & 54.2 & 91.9 & 81.9 & 67.0 & 48.9 & 50.0 & 56.2 & 79.9 \\
    HPSv3
      & 66.8 & 76.5 & 66.9 & 79.9 & 78.2 & 72.1 & 79.1 & 85.9 & 62.7 & 87.2 \\
    \midrule
    Qwen2.5-7B
      & 56.6 & 89.4 & 83.4 & 98.3 & 95.4 & 95.7 & 89.2 & 97.0 & 70.5 & 85.9 \\
    \phantom{...} HPSv3 $\Delta$ vs.\ base
      & 10.2 & -12.9 & -16.6 & -18.5 & -17.1 & -23.5 & -10.2 & -11.0 & -7.8 & 1.2 \\
    \bottomrule
    \end{tabular}
    \end{footnotesize}
  \end{center}
  \vspace{-2em}
\end{table*}

Reward models for text-to-image generation require visual understanding to assess whether a generated image faithfully depicts its conditioning prompt. 
In this section, we evaluate existing reward models on core visual reasoning competencies and find them surprisingly deficient.

\textbf{Evaluation Protocol.}
We consider three fundamental aspects of visual understanding: \textit{compositional reasoning} (recognizing multiple objects with correct attribute binding), \textit{spatial understanding} (parsing positional relationships between entities), and \textit{numeracy} (counting objects). 
These represent minimal requirements for verifying text-image alignment.
Reward models can be evaluated by their accuracy $P(x_w \succ x_l)$ over a preference pair dataset, as in Equation~\ref{eq:pref_diff_image}.
However, evaluating visual understanding through generated images introduces confounds: differences in aesthetic quality, artifacts, or stylistic variation may dominate the preference signal, obscuring whether the model can parse image content.
To isolate visual understanding, we invert the evaluation: we fix the image $x$ and assess the preference between two candidate prompts $c_w$ (correct) and $c_l$ (incorrect):
\begin{equation}
P(c_w \succ c_l \mid x) = \sigma\left( r_\phi(x, c_w) - r_\phi(x, c_l) \right)
\label{eq:pref_diff_caption}
\end{equation}
As both $c_w$ and $c_l$ are naturally formed,
this formulation localizes the preference judgment to the reward model's ability to reason visually about the correspondence between image and text.

\textbf{Benchmarks.}
We convert a diverse suite of visual understanding benchmarks in short-answer and multiple-choice format into ``preference pairs'',
where $c_w$ is assigned to the gold answer and $c_l$ to its perturbation or an incorrect multiple-choice alternative.
\textbf{Winoground} \citep{thrush2022winoground} tests compositional understanding through image-caption pairs with the same entities in different arrangements.
\textbf{SugarCrepe} \citep{hsieh2023sugarcrepe} tests basic image-text matching across three simple perturbation types: \textit{Add} (inserting spurious objects or attributes), \textit{Replace} (substituting objects or attributes), and \textit{Swap} (exchanging attributes between objects). 
\textbf{CVBench} \citep{tong2024cambrian} comprises visual-centric VQA; we convert to pairs its \textit{Relation} and \textit{Count} subsets for spatial relationships and numeracy respectively.
\textbf{WhatsUp} \citep{kamath2023s} tests spatial reasoning with both \textit{absolute} (``object in the top-left'') and \textit{relative} (``A is above B'') descriptions. 
\textbf{CountBench} \citep{paiss2023teaching} tests counting ability up to ten.
We also report accuracy on the \textbf{ImageRewardDB} test set \citep{xu2023imagereward}, to assess performance on a representative preference dataset.

\textbf{Findings.}
Table~\ref{tab:acc_rms} reports the accuracy of widely-used reward models on this benchmark suite, revealing striking deficiencies in basic visual comprehension. 
CLIP, ImageReward, PickScore, and HPSv2 perform near chance on several tasks: Winoground for compositional reasoning, CV-Relation and WhatsUp for spatial understanding, and CV-Count for numeracy. 
While HPSv3 -- a much larger model trained on more preference pairs -- generally performs better, it is far from saturating performance (66.9\% on Winoground, 62.7\% on CV-Count)
and on SugarCrepe exhibits inverse scaling relative to earlier reward models.
These failures are particularly striking given the simplicity of the underlying tasks.


Since the bag-of-words effect is well documented for text-vision encoders \citep{yuksekgonul2022and},
a natural question is whether instead using representations from a modern VLM trained on visual reasoning might resolve these deficiencies.
Indeed, HPSv3 is fine-tuned on preference data from a strong base model (Qwen2.5-VL-7B \citep{bai2025qwen25vl}), raising the question of whether its visual understanding limitations stem from the base model or the fine-tuning process. 
To investigate, we evaluate the base VLM directly as a reward model by:
\begin{align}
r(x, c) = P_\text{VLM}(\text{Yes} \mid x, 
\text{Is this a good image of: } c\text{?})
\label{eq:vlm_reward}
\end{align}
Table~\ref{tab:acc_rms} shows that Qwen2.5-VL-7B achieves strong visual understanding---83.4\% on Winoground, 97.0\% on WhatsUp, and over 95\% on all SugarCrepe subsets. After preference fine-tuning, HPSv3 gains approximately 10 percentage points in prediction accuracy on ImageRewardDB, but at severe cost: performance drops 16.6 points on Winoground, 17-24 points across SugarCrepe subsets, and 11 points on WhatsUp. The visual reasoning abilities of the base model are substantially eroded, leaving HPSv3 weaker than its foundation on the very competencies most relevant to text-image alignment.

\begin{CustomBox}{}
Popular reward models cannot reason reliably about composition, spatial relationships, and numeracy, compromising their ability to judge text-conditioned image generation.
Even starting from a VLM with strong visual understanding, these capabilities are catastrophically forgotten during preference fine-tuning.
\end{CustomBox}

\section{Robust Fine-Tuning is Not Enough to Preserve Visual Understanding}
\label{sec:continual}

\begin{wrapfigure}[19]{r}{0.55\textwidth}
\centering
\vspace{-1em}
\includegraphics[width=0.54\textwidth]{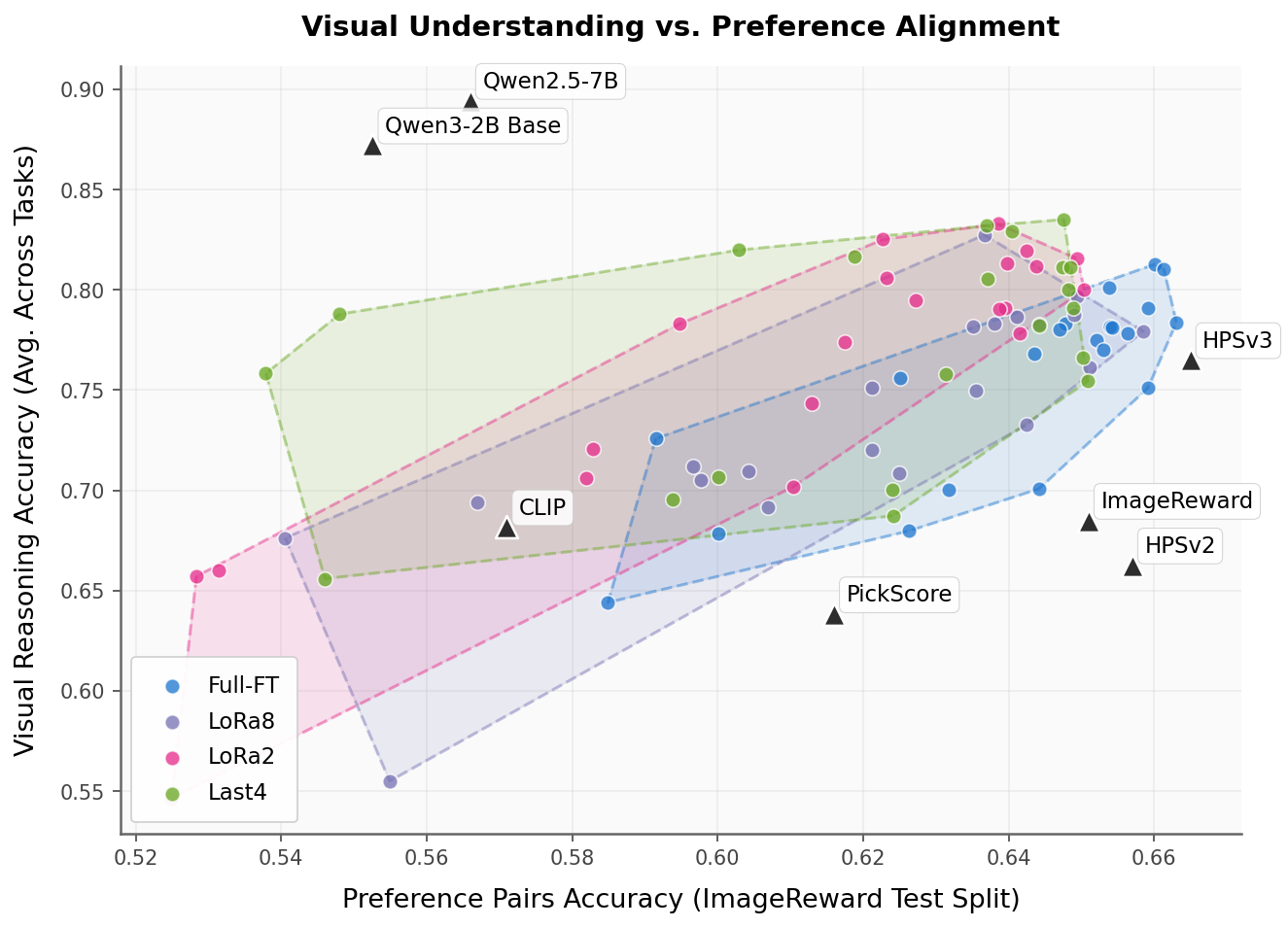}
\caption{
Visual understanding ($y$) vs preference accuracy ($x$) for multiple checkpoints of the reward model during conservative fine-tuning on preference pairs.
Conservative fine-tuning only weakly mitigates visual understanding degradation.
}
\label{fig:pareto_conservativeft}
\end{wrapfigure}

The prior section raises the question:
how can we retain a VLM's visual reasoning during preference learning?
The continual learning literature suggests possible remedies,
but we find robust fine-tuning techniques \textit{do not} adequately balance the existing visual reasoning with the new task of preference prediction.
We fine-tune Qwen3-2B \citep{bai2025qwen3} on ImageRewardDB; see Appendix~\ref{appendix:experiment_details} for details.

\subsection{Conservative Fine-Tuning}
\label{sec:conservative}

Aggressive fine-tuning induces catastrophic forgetting while more conservative updates can better preserve the original model's knowledge \citep{xuhong2018explicit,kirkpatrick2017overcoming,vu2022overcoming}. 
Thus, we investigate whether parameter-efficient fine-tuning can mitigate forgetting by regularizing towards the pretrained initialization.
We study Low-Rank Adaptation (LoRA) \citep{hu2022lora}, 
which potentially ``forgets less'' \citep{biderman2024lora};
and fine-tuning fewer layers,
since earlier layers encode more general features suitable for transfer learning \citep{neyshabur2020being}.

\textbf{Findings.} 
Figure~\ref{fig:pareto_conservativeft} shows that conservative fine-tuning offers only marginal improvements in visual understanding ($y$ axis) and at the cost of preference accuracy ($x$ axis). 
Relative to full fine-tuning, Rank-8 LoRA retains slightly better visual reasoning with losses in preference prediction; Rank-2 LoRA and fine-tuning only the last four layers show slightly stronger versions of this tradeoff. 
However, none of these methods approach the visual understanding of the base Qwen3-2B. 
Earlier checkpoints actually demonstrate worse visual understanding, suggesting the forgetting is not simply overfitting—even the most conservative configurations exhibit substantial degradation.


\begin{CustomBox}{}
Conservative fine-tuning does not adequately mitigate the erosion of visual understanding capabilities during preference learning.
\end{CustomBox}

\subsection{Weight Merging}
\label{sec:weight_merging}

Model soups, which merge the parameters of multiple fine-tuned models, can combine the strengths of different models \citep{wortsman2022robust}.
The method's theoretical foundation rests on the \textit{linear mode connectivity} (LMC) hypothesis \citep{frankle2020linear}, 
which posits that two sets of weights $\phi_1$ and $\phi_2$ derived from a shared initialization lie in the same loss basin, such that their linear interpolations satisfy
\begin{equation}
\text{Acc}\left((1-\lambda)\phi_1 + \lambda\phi_2, \mathcal{D}_{\text{test}}\right) 
\geq
(1-\lambda) \cdot \text{Acc}(\phi_1, \mathcal{D}_{\text{test}}) + \lambda \cdot \text{Acc}(\phi_2, \mathcal{D}_{\text{test}})
\label{eq:lmc}
\end{equation}
for all $\lambda \in [0, 1]$.
Weight merging with the pretrained initialization has been shown to mitigate catastrophic forgetting and improve OOD robustness across diverse continual learning settings: CLIP \citep{wortsman2022robust}, LLMs \citep{marczak2024magmax}, VLA policies \citep{yadav2025robust}, and multi-objective alignment \citep{rame2023rewarded}.
We investigate merging our preference-fine-tuned reward model and the base VLM backbone
$\phi_{\text{merged}} = (1 - \lambda) \cdot \phi_{\text{base}} + \lambda \cdot \phi_{\text{ft}}$ as a means to preserve visual reasoning capabilities.

\begin{wrapfigure}{r}{0.55\textwidth}
\centering
\vspace{-1em}
\includegraphics[width=0.54\textwidth]{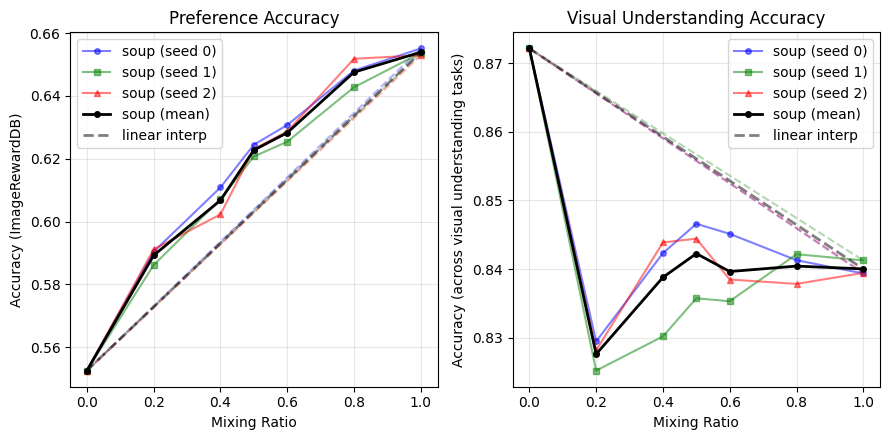}
\caption{
Preference accuracy (left) and visual understanding accuracy (right) as a function of the interpolation ratio $\lambda$ for weight merging.
Merging does not improve performance over the base ($\lambda = 0$) or fully fine-tuned ($\lambda = 1$) model. 
}
\label{fig:soup_lambda}
\end{wrapfigure}

\textbf{Findings.} 
Figure~\ref{fig:soup_lambda} shows the reward model's accuracy on ImageRewardDB and our visual understanding evaluations as a function of the interpolation coefficient $\lambda$. For preference pair accuracy, we observe the concave curve characteristic of LMC: intermediate mixing ratios outperform the linear interpolation. 
However, the best preference accuracy is still achieved by the unmerged fine-tuned model at $\lambda=1$. 
Moreover, weight merging fails to improve visual understanding. 
LMC is a claim about the loss landscape of the fine-tuning objective, and the concave curve on ImageRewardDB is the in-task behavior it anticipates. Our visual understanding suite is an off-task evaluation that LMC does not constrain, and although merging can improve OOD robustness in practice \citep{wortsman2022robust, rame2023ratatouille}, that transfer does not occur here: the merged model's visual understanding falls below the linear interpolation between endpoints and is especially poor at small mixing ratios.

\begin{CustomBox}{}
Weight merging does not improve over the fine-tuned model for either preference learning or visual reasoning retention, and in the latter case does not exhibit LMC.
\end{CustomBox}

\subsection{Calibrating VLM Uncertainty}
\label{sec:calibrate}

Rather than training a regression head to extract rewards from the VLM backbone, an alternative approach is to directly calibrate the model's existing output distribution. 
Equation~\ref{eq:vlm_reward} extracts the probability of the VLM generating ``Yes'' to a reward query; we fine-tune these probabilities to align with the preference dataset.
This approach occupies an intermediate position between discriminative and generative reward modeling paradigms. 
Discriminative reward models predict fine-grained continuous scores typically learned with the Bradley-Terry objective. 
Generative reward models, in contrast, prompt LLMs (or VLMs) to produce verbalized judgments \citep{zhang2024generative,wang2024self,mahan2024generative},
which has been hypothesized to better tap into their base reasoning capabilities by leveraging the model's native next-token prediction interface \citep{zhang2024generative,wu2025rewarddance}.
Logit calibration could potentially inherit advantages from both paradigms: operating through next-token-prediction yet still producing continuous rewards suitable for fine-grained signals. 
We use the Bradley-Terry objective to directly adjust the VLM's confidence estimates:
\begin{align}
&\mathcal{L}(\phi) = -\mathbb{E}_{(c, x_w, x_l)} \big[ \log \sigma \big( P_\phi(\text{``Yes''} \mid x_w, \text{``Is this a good image of: } c\text{?''}) - 
\nonumber \\
& \phantom{........................} P_\phi(\text{``Yes''} \mid x_l, \text{``Is this a good image of: } c\text{?''}) \big) \big].
\end{align}

\begin{wrapfigure}[18]{r}{0.55\textwidth}
  \centering
  \vspace{-1em}
  \includegraphics[width=0.54\textwidth]{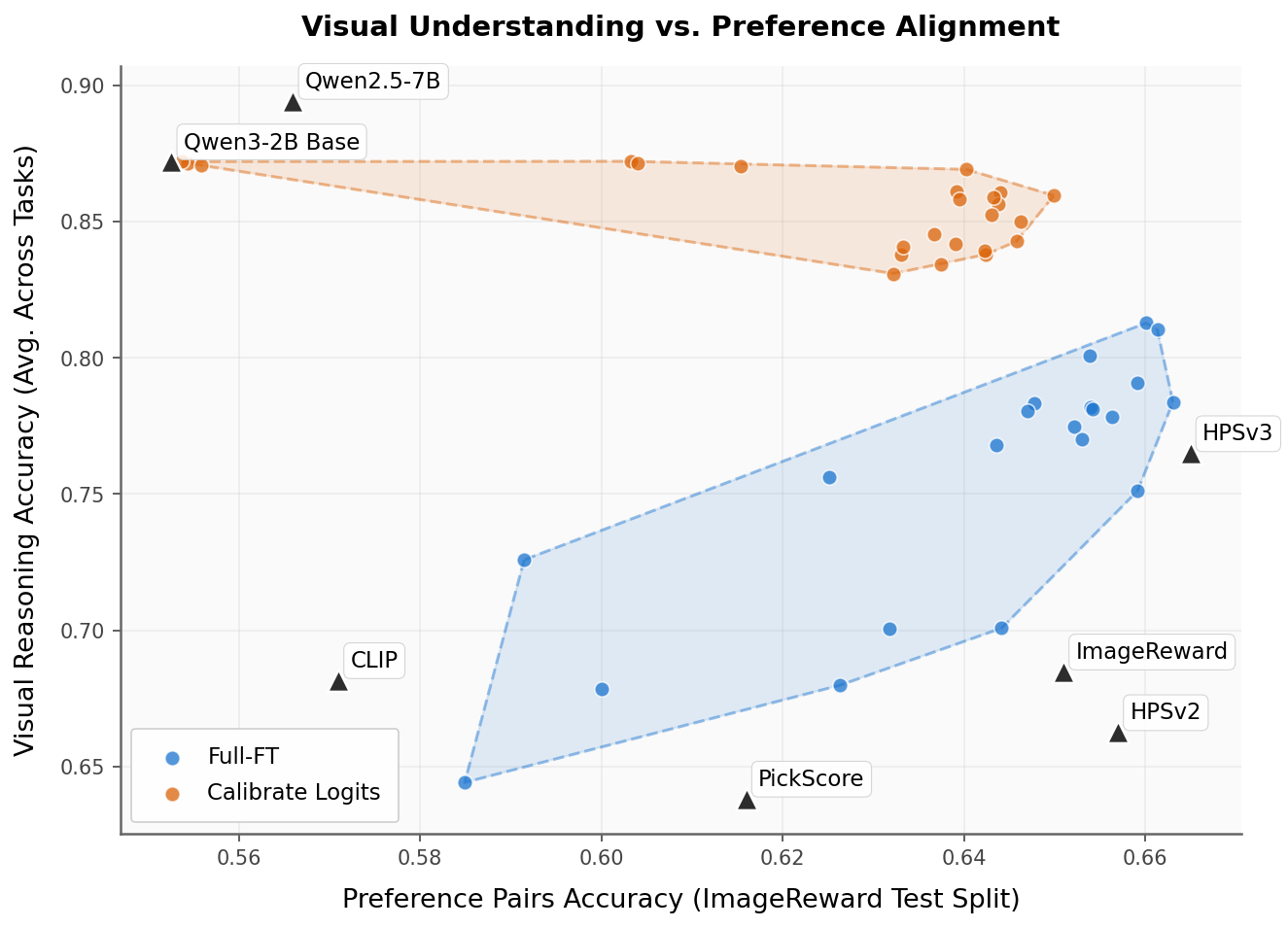}
  \caption{
  Visual understanding ($y$) vs preference accuracy ($x$) for multiple checkpoints of the reward model trained with uncertainty calibration (orange) and with the standard reward readout head (blue).
  Uncertainty calibrating the base VLM better retains its visual reasoning, but underperforms on preference prediction.}
  \label{fig:pareto_calibrate}
  \vspace{-1em}
\end{wrapfigure}

\textbf{Findings.} 
Uncertainty calibration better preserves visual reasoning but underfits preferences, making it a weak candidate for the preference learning objective.
Figure~\ref{fig:pareto_calibrate} shows that at early checkpoints, 
visual understanding remains near the base model's level, and decays only mildly as training progresses,
remaining substantially above the regression head baseline throughout. 
This is consistent with the hypothesis that operating through the VLM's existing generative interface better preserves pretrained representations. 
However, this preservation comes at a cost: the constrained optimization limits capacity to fit human preferences, underperforming the regression head standard by 2 percentage points.

\begin{CustomBox}{}
Calibrating VLM uncertainty underfits the preference dataset, though it largely preserves visual reasoning. 
\end{CustomBox}

\section{CrossPairs: A Data-Centric Solution}
\label{sec:datacentric}

Standard preference datasets pair a single prompt with two images of differing quality. The learning signal from such pairs conflates two distinct objectives: discerning aesthetic quality and assessing text-image alignment. When both images differ in style and artifacts alongside their semantic content, the model can exploit aesthetic shortcuts rather than developing robust visual parsing. Simply scaling preference datasets is not an effective remedy: the progression from 130k pairs in ImageRewardDB to 1.2M in HPSv3 still produces reward models with suboptimal visual understanding (Section~\ref{sec:diagnosis}), and can even lead reward models to overindex on stylistic properties rather than genuine image comprehension, as we observe in Figure~\ref{fig:examples}. To target visual reasoning directly, we need preference pairs whose learning signal better isolates text-image alignment.

We propose \textit{CrossPairs}: preference pairs that share a common image but differ in prompt—one correct description and one incorrect. By holding the image fixed, aesthetic confounds are eliminated and the learning signal targets text-image alignment directly. 
Conveniently, this structure maps naturally onto existing VQA and multiple-choice data for vision-centric multimodal learning,
allowing us to leverage existing training data without generating new images or collecting human annotations.
These pairs are mixed with the standard shared-prompt pairs during preference learning.
We used the CrossPairs structure to \textit{evaluate} visual reasoning in Section~\ref{sec:diagnosis}; here we leverage it for \textit{training}.

We offer some intuition for how CrossPairs isolate visual reasoning.
Consider a reward model $r_\phi(x, c)$ that assigns a scalar score to an image-prompt pair. 
We can define an image-only component of this reward that captures prompt-independent aesthetic properties, by marginalizing over the distribution over prompts $\mathcal{C}$ :
\begin{equation}
r_\phi^{x}(x) := \mathbb{E}_{c \in \mathcal{C}} [r_\phi(x, c)]
\end{equation}
Then, we can define the residual
\begin{equation}
\tilde{r}_\phi(x,c) = r_\phi(x,c) - r_\phi^{x}(x),
\end{equation}
as a term that isolates the reward signal pertaining to prompt-image alignment.

For the standard shared-prompt preference pair $(c,x_w,x_l)$,
a gradient step with Bradley-Terry loss yields the update
\begin{align}
&- \nabla_\phi \log \sigma [ r_\phi(x_w, c) - r_\phi(x_l, c) ] 
= - \nabla_\phi \log \sigma \bigg[ \big( r_\phi^{x}(x_w) + \tilde{r}_\phi(x_w, c) \big) -
\big( r_\phi^{x}(x_l) + \tilde{r}_\phi(x_l, c) \big) \bigg]
\end{align}
This decomposition reveals that even when prompts $c$ are carefully selected to target compositional, spatial, or numeracy skills,
painstaking quality control is necessary to ensure that $x_w$ and $x_l$ differ on correctness and not image aesthetic quality.
In the above loss, it is possible to learn shortcut correlates rather than develop robust visual reasoning.

In contrast, a CrossPair is a triple $(c_w, c_l, x)$: a common image $x$ with a preferred caption $c_w$ and a dispreferred caption $c_l$.
The Bradley-Terry gradient update becomes:
\begin{align}
&- \nabla_\phi \log \sigma \bigg[ \big( r_\phi^x(x) + \tilde{r}_\phi(x, c_w) \big) - 
\big( r_\phi^x(x) + \tilde{r}_\phi(x, c_l) \big) \bigg]
= - \nabla_\phi \log \sigma \big[ \tilde{r}_\phi(x, c_w) - \tilde{r}_\phi(x, c_l) \big]
\end{align}
The gradient $\nabla \mathcal{L}$ thus depends only on how the reward model discriminates between prompts of the same image, 
which isolates visual understanding capability we wish to train.
This optimization is invariant to any additive reparametrization of the residualized image-only component, $r_x \mapsto r_x + g(x) $ for arbitrary $g$.

\begin{wraptable}{r}{0.38\textwidth}
  \vspace{-1em}
  \caption{Final data mix for reward model training, comprising ImageRewardDB and multimodal learning VQA converted to CrossPairs preference pairs.}
  \label{tab:data_mix}
  \begin{center}
    \begin{small}
      \begin{tabular}{lrr}
        \toprule
        Dataset  & Pairs & \% Mix \\
        \midrule
        CLEVR    & 8750 & 5.6 \\
        SpatialSense & 9712 & 6.2 \\
        TallyQA & 10318 & 6.6 \\
        ColorSwap & 2000 & 1.3 \\
        VSR & 6664 & 4.2 \\
        ImageRewardDB & 119765 & 76.2 \\
        \midrule
        Total & 157209 & 100.0 \\
        \bottomrule
      \end{tabular}
    \end{small}
  \end{center}
\end{wraptable}

\textbf{Data.} We convert diverse VQA datasets into CrossPairs format by pairing correct captions with targeted perturbations. 
We leverage the following datasets.
\textbf{CLEVR} \citep{johnson2017clevr} comprises Blender-rendered scenes of geometric primitives 
with varying colors, sizes, and materials. 
This dataset targets compositional understanding and attribute binding.
We construct negative captions by swapping attributes between objects in the scene.
\textbf{SpatialSense} \citep{yang2019spatialsense} contains real images from Flickr 
and \textbf{VSR} \citep{liu2023visual} contains natural photographs from MS COCO.
Both of these are annotated with spatial relation statements, and we form the negative captions by inverting the relations that have a meaningful opposite (e.g. left/right, above/under).
\textbf{TallyQA} \citep{acharya2019tallyqa} focuses on counting in realistic, complex scenes from Visual Genome \citep{krishna2017visual}. 
We rephrase its VQA format into captions, 
and create negative captions by perturbing the true answer by $\pm1$.
\textbf{ColorSwap} \citep{burapacheep2024colorswap} contains AI-generated images targeting color-object binding,
where pairs of prompts (and their corresponding images) differ only in color-object binding. 
Negative prompts are the alternative in the pair (e.g., ``a black mug with red liquid'' $\leftrightarrow$ ``a red mug with black liquid''). 
Table~\ref{tab:data_mix} shows the resultant training mixture comprising the shared-image CrossPairs along with ImageRewardDB \citep{xu2023imagereward}.

\subsection{Results}
\label{sec:datacentric_results}

We train Qwen3-2B and a small reward readout head on the final CrossPairs mix (Table~\ref{tab:data_mix}) for three epochs.
We first evaluate reward model accuracy directly, then assess the quality of generations guided by our reward model in both inference-time scaling and reinforcement learning settings. We also add VQAScore \citep{lin2024evaluating} and UnifiedReward \citep{wang2025unified} as baselines representing more recent, ``reasoning-oriented'' reward models.


\textbf{Pareto improvement on preference prediction.} 
Figure~\ref{fig:fig1} shows that our CrossPairs data intervention achieves a Pareto improvement on the joint objective of preference accuracy and visual understanding. Our method achieves 66.3\% accuracy on ImageRewardDB -- comparable to the best reward baseline -- while maintaining much stronger visual understanding than all reward models that have been calibrated on human preferences. 
(See Appendix~\ref{appendix:rm_accuracies} for a breakdown.)
Our method is the only one to reach the top-right region of the plot, while continual learning approaches and existing reward models exhibit suboptimal tradeoffs.
We conduct a controlled comparison of our reward model, ablating out the inclusion of CrossPairs (Appendix~\ref{appendix:ablating_out_crosspairs}) and show that between reward models trained on the same preference dataset, adding CrossPairs supervision preserves aesthetic quality (Appendix~\ref{appendix:aesthetic}).
Next, we systematically assess our reward model by measuring the quality of text-to-image generations it guides.


\begin{figure}[t]
    \centering
    \begin{subfigure}[t]{0.495\columnwidth}
        \centering
        \includegraphics[width=\linewidth]{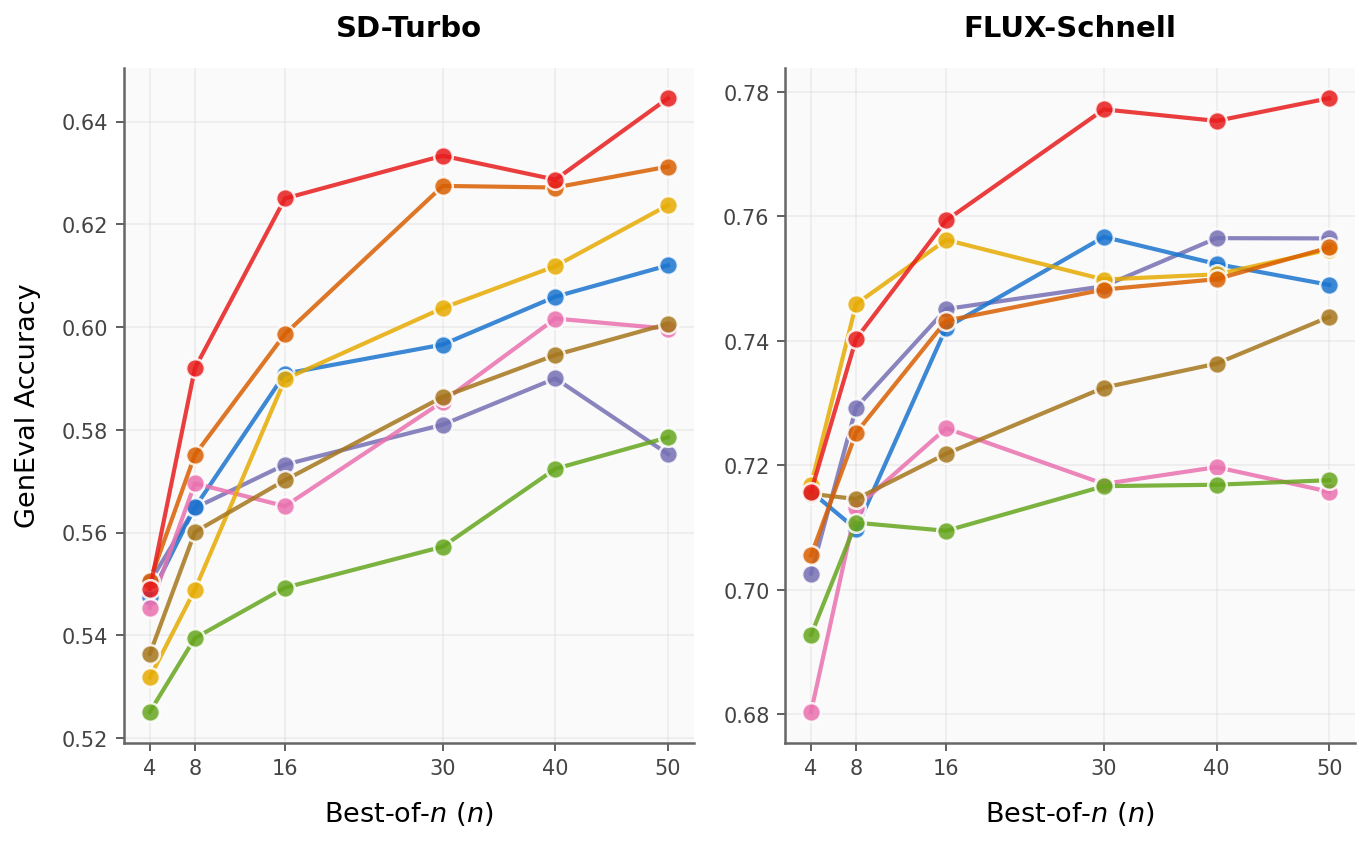}
        \caption{Best-of-$n$ filtering.}
        \label{fig:geneval_bon}
    \end{subfigure}
    \hfill
    \begin{subfigure}[t]{0.495\columnwidth}
        \centering
        \includegraphics[width=\linewidth]{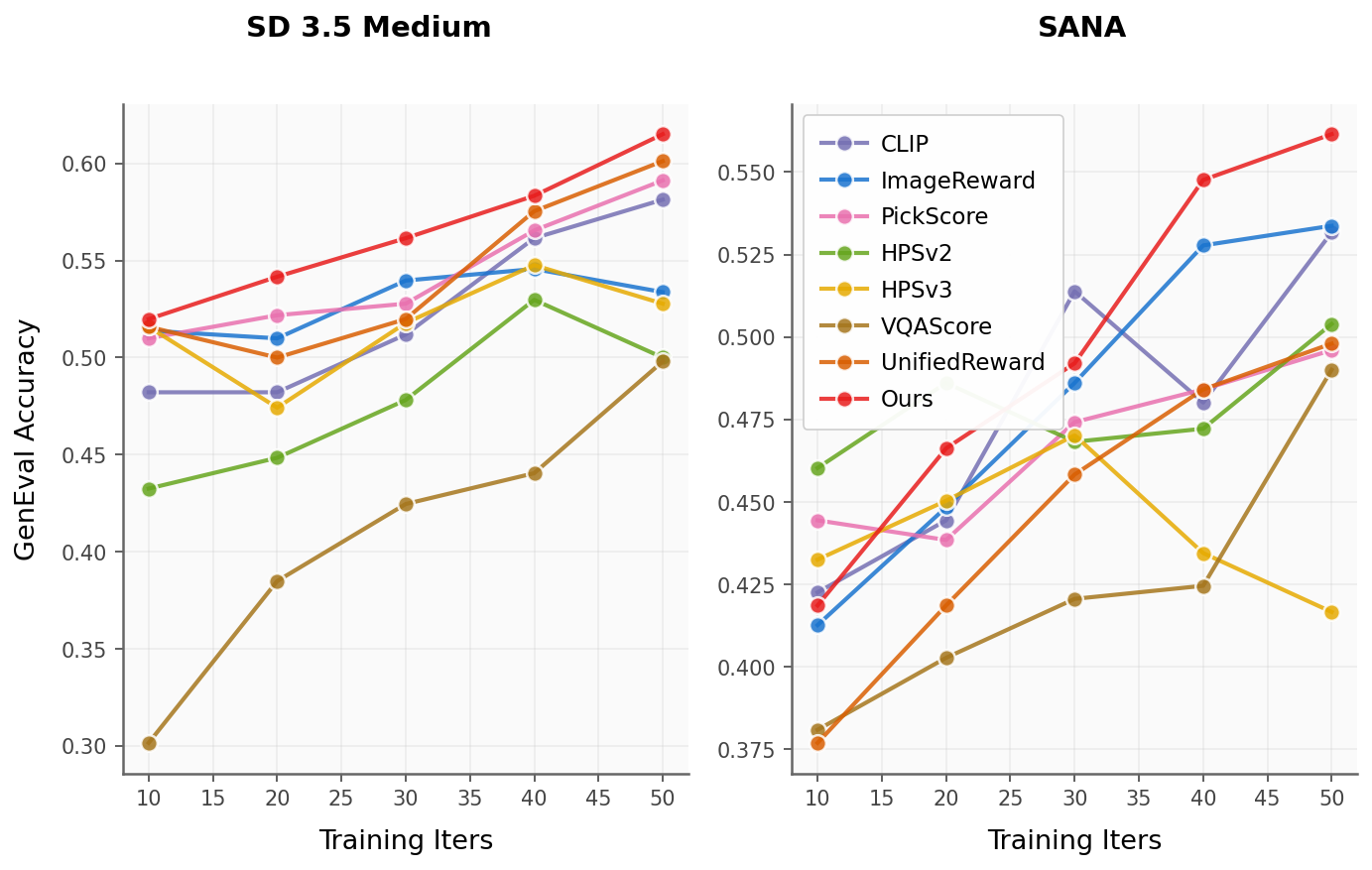}
        \caption{RL with DiffusionNFT.}
        \label{fig:geneval_rl}
    \end{subfigure}
    \caption{Our reward model improves both inference-time scaling (left: single-step SD-Turbo and FLUX.1-Schnell) and post-RL performance (right: SD 3.5 Medium and SANA) on GenEval.}
    \label{fig:combined_geneval}
    \vspace{-2em}
\end{figure}

\textbf{Inference-time scaling.} Our reward model yields the strongest inference-time scaling on challenging compositional generation.
We evaluate best-of-$n$ selection, a simple yet strong baseline for improving generation quality without weight updates \citep{ma2025inference},
on GenEval \citep{ghosh2023geneval}, which explicitly tests compositional prompts requiring precise rendering of objects, attributes, spatial relations, and quantities.
Figure~\ref{fig:geneval_bon} shows GenEval accuracy as a function of $n$ for two single-step generators, SD-Turbo \citep{sauer2024adversarial} and FLUX.1-Schnell \citep{flux2024}. 
Our reward model consistently outperforms all baselines across the curve as $n$ scales. 
On SD-Turbo, our reward model achieves 64.5\% accuracy at $n=50$, compared to 62.4\% for the next-best method. 
On FLUX.1-Schnell, we also see pronounced gains with our reward model reaching 78.1\% accuracy with $n=50$, compared to the next best at 75.7\%. 
Our reward model outperforms all baselines on both a lightweight and more capable generator, indicating that it offers the most reliable signal for correctly rendering difficult compositional prompts.
See Appendix~\ref{appendix:further_generators} for extended results with other generators.

\textbf{Reinforcement learning.}
Our reward model provides a more effective signal for RL post-training.
We use DiffusionNFT \citep{zheng2025diffusionnft}, a recent state-of-the-art on-policy RL algorithm for diffusion,
to train Stable Diffusion 3.5 Medium \citep{stabilityai2024sd35} and SANA \citep{xie2024sana} with prompts from Pick-a-Pic.
Figure~\ref{fig:geneval_rl} shows GenEval accuracy across training iterations. 
Our reward model yields steady improvement throughout training, reaching 61.5\% (SD) and accuracy at 50 iterations, 
and dominates other reward models across the training curve. 
Notably, GenEval is out-of-distribution relative to the Pick-a-Pic distribution used for the RL steps, yet consistent improvement under our reward model suggests it provides a more reliable signal for learning compositional generation, 
grounded in its strong visual understanding.

\begin{figure}
\centering
\includegraphics[width=\textwidth]{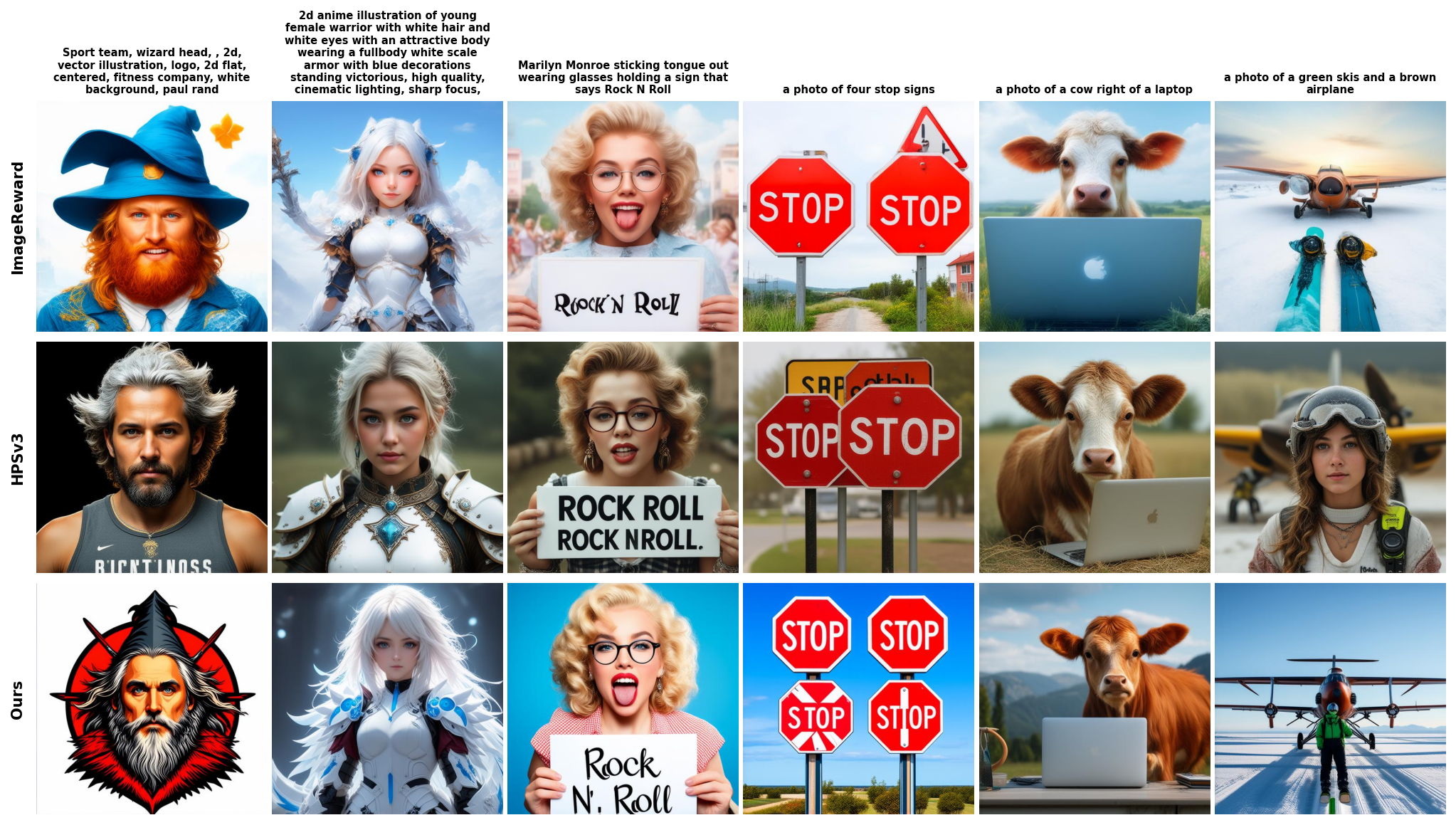}
\caption{Generated samples after post-training with different reward models. The first three columns are prompts from Pick-a-Pic; the last three columns are prompts from GenEval.}
\vspace{-1.5em}
\label{fig:examples}
\end{figure}

\textbf{Qualitative analysis.} 
Figure~\ref{fig:examples} illustrates that our reward model guides generations with both strong aesthetics and strong prompt adherence.
ImageReward-guided generations follow prompts reasonably well but exhibit somewhat weaker aesthetics: images appear less sharp and with less polished overall composition.
Both HPSv3 and our reward model trained with CrossPairs produce crisp and visually appealing images. 
However, HPSv3 tends toward photorealistic, conventionally attractive faces regardless of prompt. 
For example, the HPSv3-post-trained generative model ignores the prompt specifications ``2d, vector illustration, logo... white background'', producing a photorealistic portrait on a black background (first column). 
The last three columns, drawn from GenEval's compositional prompts, illustrate the qualitative outcomes of our reward model's stronger visual understanding, corroborating Figure~\ref{fig:geneval_rl}. 
These prompts remain challenging even after post-training. 
For example, for ``a photo of four stop signs'', 
our reward model is the only one to produce the correct count, despite some distortions.
See Appendix~\ref{appendix:further_examples} for further examples across all reward models.

Our results support the core claim: 
visual understanding is essential to a strong reward model, 
but simply scaling up preference pair annotations on the natural distribution of user prompts fails to target the visual reasoning required to assess complex compositions.
Our approach yields a reward model that improves text-image alignment with strong aesthetics.

\section{Related Work}

\textbf{Reward models for image generation.}
Reward models are central to aligning text-to-image models with human preferences. 
Early work leveraged dual text-vision encoders like CLIP \citep{radford2021learning} and BLIP \citep{li2022blip} as a zero-shot reward signal;
subsequent work fine-tuned them on human preference datasets  \citep{xu2023imagereward,kirstain2023pick,wu2023hpsv1, wu2023humanv2,zhang2024learning}.
More recently, VLMs have been explored as reward model backbones: 
VQAScore \citep{lin2024evaluating} extracts rewards from response probabilities, 
HPSv3 \citep{ma2025hpsv3} fine-tunes VLMs on preference data,
LLM-as-judge approaches prompt for verbal assessments \citep{cho2023davidsonian,ku2024viescore},
and generative reward models produce judgments in natural language \citep{wang2025unified}.
Our work studies the tension between leveraging VLMs' visual understanding and fitting to preferences, finding that standard fine-tuning erodes the capabilities that make VLMs attractive as backbones.

\textbf{Fine-tuning VLMs.}
VLMs acquire broad capabilities through web-scale pretraining, but fine-tuning often causes catastrophic forgetting \citep{wang2024comprehensive, luo2025empirical}, particularly in VLMs due to cross-modal feature drift and zero-shot erosion \citep{liu2025continual}. Parameter-efficient methods like LoRA \citep{biderman2024lora} and weight merging \citep{ marczak2024magmax,sokar2025continual} may offer partial remedies. Similar degradation occurs when fine-tuning VLMs for robotic control, where forgetting overwrites visual-text alignments \citep{dey2025revla,zhou2025chatvla,hancock2025actions}. Our work shows preference fine-tuning erodes visual reasoning in reward models.

\textbf{Robust reward learning.} 
Human preference annotations exhibit noise from annotator disagreement and labeling errors \citep{bai2022training,ouyang2022training}, with inter-annotator agreement as low as 73\% \citep{shen2024towards}. Techniques to address this include label smoothing \citep{zhu2024iterative}, adaptive margins and label flipping \citep{wang2024secrets}, and debiased DPO \citep{chowdhury2024provably}. 
These methods target noise in labels; our work addresses a complementary problem — inadequate reward signal for visual grounding in text-to-image reward models.

\section{Conclusion}

Visual understanding is essential to reward modeling, yet preference fine-tuning erodes these capabilities, even when starting from strong VLM backbones. 
By injecting visual reasoning supervision directly into preference learning using CrossPairs, we achieve a Pareto improvement over existing reward models.
Our reward model guides both inference-time selection and RL post-training toward aesthetic generations with stronger prompt adherence.

\begin{ack}

S.S. was supported by the Stanford School of Engineering fellowship and by research funding to the Yeung-Levy lab. Compute was provided by the Stanford AI Lab (SAIL) cluster. This work did not receive additional third-party funding and the authors declare no competing interests.

\end{ack}

\clearpage
\bibliographystyle{plainnat}
\bibliography{references}

\FloatBarrier

\clearpage
\appendix
\section{Extended Preliminaries}
\label{appendix:prelims}

\subsection{Reward Models in Text-to-Image Generation}
\label{sec:prelims_usage}
Reward models play a central role in aligning text-to-image generative models with human preferences, appearing in both inference-time and post-training pipelines.

\textbf{Inference-time scaling.} Reward models can be leveraged at inference time to improve generation quality without modifying model weights. The simplest approach is best-of-$n$ sampling on the final generation: given a prompt $c$, we generate $n$ candidate images $\{x_1, \ldots, x_n\}$ and return the sample with highest reward:
\begin{equation}
x^* = \arg\max_{x_i} \; r_\phi(x_i, c)
\end{equation}
Beyond simple reranking, search-based methods explore the space of denoising trajectories using techniques such as beam search, particle filtering, and rejection sampling to identify higher-quality paths \citep{ma2025inference, liu2024correcting, na2024diffusion, karthik2023if}. 
Alternatively, the initial noise or intermediate latents can be directly optimized 
via gradient-based methods to maximize reward signals \citep{eyring2024reno, wallace2023end}. 
Reward gradients can also be incorporated into the score function at each 
denoising step, steering sampling toward higher-reward regions \citep{dhariwal2021diffusion, ho2022classifier,bansal2023universal}.

\textbf{Post-training with reinforcement learning.}
Given a pretrained diffusion or flow-based generative model, RL can optimize the model to produce images that receive high reward. 
Let $p_\theta(x|c)$ denote the distribution over images $x$ induced by a generative model with parameters
$\theta$ conditioned on prompt $c$, and let $r_\phi(x,c)$ denote a reward model. 
In broad strokes, the RL objective maximizes expected reward while regularizing against a reference model $p_{\text{ref}}$:
\begin{equation}
\max_\theta \; \mathbb{E}_{c \sim \mathcal{D}, x \sim p_\theta(\cdot | c)} \left[ r_\phi(x, c) \right] - \beta \, \text{KL}\left( p_\theta(\cdot | c) \| p_{\text{ref}}(\cdot | c) \right)
\end{equation}
where $\beta$ controls the strength of the regularization. 
Many methods have been proposed to optimize this objective for diffusion models \citep{black2023training, fan2023dpok, clark2023directly, liu2025flow, zheng2025diffusionnft};
while their algorithms differ, all rely on a reward signal to guide optimization.

\subsection{Reward Modeling with Bradley-Terry}
\label{sec:prelims_bt}

\textbf{Preference pairs.} Text-to-image reward models are typically trained on datasets of pairwise human preferences. Given a prompt $c$ and two generated images $x_w$ (preferred) and $x_l$ (dispreferred), the training data consists of tuples $(c, x_w, x_l)$ indicating that human annotators preferred $x_w$ over $x_l$ as a response to $c$.

\textbf{Bradley-Terry model.} 
We assume the Bradley-Terry model \citep{bradley1952rank} for human preferences.
Let $r_\phi(x, c)$ denotes the scalar reward assigned to image $x$ given prompt $c$, the probability that $x_w$ is preferred over $x_l$ is modeled as:
\begin{equation}
P(x_w \succ x_l \mid c) = \sigma\left( r_\phi(x_w, c) - r_\phi(x_l, c) \right)
\end{equation}
where $\sigma(\cdot)$ denotes the sigmoid function. This induces a maximum likelihood training objective over a dataset $\mathcal{D}$ of preference pairs:
\begin{equation}
\mathcal{L}(\phi) = -\mathbb{E}_{(c, x_w, x_l) \sim \mathcal{D}} \left[ \log \sigma\left( r_\phi(x_w, c) - r_\phi(x_l, c) \right) \right]
\end{equation}
This formulation posits that preferences depend only on the individual qualities of the two images and not on interactions between them—an independence assumption that enables pointwise reward modeling.
In this work, we focus on \textit{pointwise} reward models that produce an absolute quality assessment $r_\phi(x, c) \in \mathbb{R}$ for a prompt-image pair, independent of other candidate images. 
They contrast with pairwise reward models, which require a reference image for comparison and cannot score candidates in isolation. 
They also contrast generative reward models, which produce rewards in natural language and are therefore typically limited to binary judgments or discrete rating scales rather than fine-grained continuous scores \citep{zhang2024generative, zhang2023gpt}. 
Pointwise models offer practical advantages for deployment: they provide absolute quality assessments suitable for thresholding, filtering, and reference-free RL.

\section{Experiment Details}
\label{appendix:experiment_details}

\subsection{Experiment details for Section~\ref{sec:continual}}

To isolate the effects of training methodology from confounding factors, we hold fixed the base model (Qwen3-2B) and train for 3 epochs on ImageRewardDB throughout Section~\ref{sec:continual}. 
We use Qwen3-2B as our base VLM, a capable model that balances strong visual understanding with computational tractability. 
For training data, we use ImageRewardDB \citep{xu2023imagereward}, from which we extract only the pairwise rank comparisons (discarding absolute ratings) to match the Bradley-Terry formulation.
Unless otherwise stated, our baseline configuration performs full fine-tuning on the last 9 layers of the language model (comprising approximately one-third of the LLM parameters) of the VLM, with the visual encoder and adapter fully frozen.
We find this configuration sufficient to match state-of-the-art performance on the ImageRewardDB held-out test set while providing a strong baseline for investigating forgetting mitigation strategies.

\subsection{Experiment details for Section~\ref{sec:conservative}}

We experiment with conservative fine-tuning.
In addition to the baseline of full fine-tuning on the last 9 layers,
we fine-tune those same layers with 1) rank 8 LoRA (with $\alpha=16$) and 2) rank 2 LoRA (with $\alpha=4$), and 3) also perform full-finetuning on the last 4 layers.
In all cases, given an image $x$ and prompt $c$, the VLM produces a joint representation $h=f_\theta(x, c)$ on the last token, 
to which we attach a regression head outputting a scalar reward: $r_\phi(x,c)=w^T h + b$. 
This lightweight linear layer is randomly initialized and always fully-finetuned.
For each configuration, we train with 3 randomly seeded data orders and save 8 checkpoints throughout training, yielding 24 model snapshots.

\subsection{Experiment details for Section~\ref{sec:weight_merging}}

We experiment with weight merging.
We first train only the readout head for 1 epoch on ImageRewardDB before unfreezing the last 9 LLM layers of Qwen3-2B for 3 epochs of end-to-end full fine-tuning.
This two-stage strategy helps to keep the fine-tuned model in the same loss basin as the initialization \citep{kumar2022fine,rame2024warm}.
For model merging, we interpolate the VLM backbone with coefficient $\lambda$ by $\phi_{\text{merged}} = (1 - \lambda) \cdot \phi_{\text{base}} + \lambda \cdot \phi_{\text{ft}}$.
The readout head parameters are taken from the fine-tuned reward model (the base VLM has no corresponding component). 

\subsection{Experiment details for Section~\ref{sec:calibrate}}

We experiment with uncertainty calibration of the VLM's confidence estimates. 
Without preference fine-tuning, out-of-the-box VLMs produce a probability $P(\text{"Yes"})$ reflecting its assessment about image-text alignment, but this probability is poorly calibrated with respect to human preferences (Table~\ref{tab:acc_rms}). 
We use the Bradley-Terry objective to directly adjusts the VLM's confidence estimates:
\begin{align}
&\mathcal{L}(\phi) = -\mathbb{E}_{(c, x_w, x_l)} \big[ \log \sigma \big( 
\nonumber \\
& \phantom{.....} P_\phi(\text{``Yes''} \mid x_w, \text{``Is this a good image of: } c\text{?''}) -
\nonumber \\
& \phantom{.....} P_\phi(\text{``Yes''} \mid x_l, \text{``Is this a good image of: } c\text{?''}) \big) \big]
\end{align}
We fine-tune the last 9 layers of the Qwen3-2B VLM, consistent with Section~\ref{sec:conservative}, and do not make use of a readout regression head.
The experiment is repeated for 3 random seeds and we save 8 checkpoints during the course of training over 3 epochs.

\subsection{Experiment details for Section~\ref{sec:datacentric_results}}

Our final reward model is obtained by fine-tuning on the last 9 layers of Qwen3-2B VLM and a linear projection head. 
We train for three epochs with Bradley-Terry loss on the combined dataset described in Table~\ref{tab:data_mix}. 
This setup is deliberately straightforward — we use default hyperparameters matching those from Section~\ref{sec:continual}, demonstrating that the performance gains arise from the data intervention rather than careful tuning.

For reinforcement learning with DiffusionNFT,
we train on prompts from the PickScore dataset using Stable Diffusion 3.5 Medium as the base model, with 10 denoising steps and no classifier-free guidance. 
These settings produce relatively poor samples from the base model, providing a challenging testbed for whether RL can unlock generation quality.
We set $\beta=1$ for DiffusionNFT
and train on 8 sampled images per prompt.

\section{Further Results}
\label{appendix:results}

\subsection{Reward Model Accuracies}
\label{appendix:rm_accuracies}

\begin{table*}[h]
  \caption{Accuracy of reward models on ImageRewardDB and visual understanding evaluations. Random performance is 50\%. $^\dagger$Trained only as a pairwise model; evaluated here by prompting for a pointwise rating (see text).}
  \label{tab:acc_rms_new}
  \begin{center}
    \begin{footnotesize}
    \setlength{\tabcolsep}{1.5pt}
    \begin{tabular}{lccrrrrrrrr}
    \toprule
      &
      \multirow{2}{*}{\centering ImageRDB} &
      \multirow{2}{*}{\centering Visual} &
      \multicolumn{8}{c}{Visual Understanding Breakdown} \\
      \cmidrule(lr){4-11}
      & & &
      Wground & SC-Add & SC-Rep & SC-Swap & CV-Rel & WhatsUp & CV-Count & CountBch \\
    \midrule
    CLIP
      & 57.1 & 68.2 & 55.4 & 89.9 & 85.4 & 71.9 & 53.5 & 48.9 & 57.8 & 82.6 \\
    ImageReward
      & 65.1 & 68.5 & 56.8 & 96.1 & 88.4 & 83.9 & 51.8 & 52.0 & 55.6 & 63.4 \\
    PickScore
      & 61.6 & 63.8 & 54.6 & 81.6 & 80.6 & 64.9 & 51.7 & 50.6 & 52.8 & 73.8 \\
    HPSv2
      & 65.7 & 66.3 & 54.2 & 91.9 & 81.9 & 67.0 & 48.9 & 50.0 & 56.2 & 79.9 \\
    HPSv3
      & 66.8 & 76.5 & 66.9 & 79.9 & 78.2 & 72.1 & 79.1 & 85.9 & 62.7 & 87.2 \\
    VQAScore
      & 56.8 & 79.2 & 72.9 & 92.3 & 88.8 & 86.1 & 79.2 & 84.6 & 63.4 & 66.0 \\
    UnifiedReward
      & 60.0 & 77.2 & 75.9 & 92.1 & 88.9 & 85.0 & 69.2 & 74.9 & 59.5 & 71.8 \\
    UnifiedReward-Think$^\dagger$
      & 53.6 & 69.6 & 66.1 & 81.8 & 77.0 & 73.0 & 63.2 & 68.8 & 63.2 & 64.0 \\
    Qwen3-2B
      & 55.3 & 85.4 & 75.9 & 98.2 & 94.2 & 93.4 & 87.3 & 91.2 & 60.4 & 82.8 \\
    Ours w/o CrossPairs
      & 66.3 & 78.4 & 66.8 & 93.7 & 88.1 & 84.4 & 78.9 & 73.2 & 59.0 & 82.9 \\
    Ours
      & 66.3 & 85.3 & 71.6 & 97.5 & 92.8 & 91.9 & 74.5 & 92.2 & 71.7 & 90.2 \\
    \bottomrule
    \end{tabular}
    \end{footnotesize}
  \end{center}
\end{table*}

Table~\ref{tab:acc_rms_new} reports the full accuracy results for all reward models on ImageRewardDB and our visual understanding benchmark suite, extending Table~\ref{tab:acc_rms}
in the main body to include VQAScore, UnifiedReward, and our ablated baseline `Ours w/o CrossPairs' (described in Section~\ref{appendix:ablating_out_crosspairs}).

VQAScore extracts a reward by querying a VLM with ``Does this figure show \{prompt\}?'' and reading off the probability of the ``Yes'' token, leveraging the VLM's pretrained visual reasoning directly without preference fine-tuning. UnifiedReward is a generative reward model that prompts a VLM to verbalize a discrete quality rating in natural language; it is trained on both pointwise ratings and pairwise preferences (both images presented to the VLM simultaneously). 

We additionally report UnifiedReward-Think \citep{wang2026unified}, which extends UnifiedReward with long-form chain-of-thought and RL-based reasoning enhancement and is therefore the closest reasoning-oriented reward model to our motivation. It differs from every other model we study in an important respect: it is trained only as a pairwise preference model, emitting a relative judgment over two images, and is not trained to rate a single image. Because the need for a reference image is unwieldy for best-of-$n$ selection and RL, we do not use it as a downstream reward signal. Since it is a generative model, however, we can prompt it for a pointwise rating; it returns a parseable score in most cases, despite occasional format errors. The resulting score is poorly calibrated: it trails even CLIP on ImageRewardDB and is only marginally above CLIP on average visual understanding, well below UnifiedReward and our model on both. Inference is also substantially slower owing to the long prompt and chain-of-thought generation.

Several observations stand out from the results of Table~\ref{tab:acc_rms_new}:

\textbf{Existing reward models trade off visual reasoning for preference accuracy.} No baseline reward model trained on human preferences achieves both strong preference prediction and strong visual reasoning. Among baselines, HPSv3 achieves the strongest balance, though still short of our method on visual understanding. Notably, HPSv3 uses substantially more preference data (1.3M pairs) and a larger backbone (7B) than our setup; we do not claim a controlled efficiency comparison, but rather note that scaling preference pairs alone has not closed the visual reasoning gap.

\textbf{Reasoning-oriented reward models are stronger visual reasoners but weaker on preferences.} VQAScore and UnifiedReward achieve relatively strong visual understanding (79.2\% and 77.2\% average) but underperform on preference prediction (56.8\% and 60.0\% on ImageRewardDB). For VQAScore, this likely reflects that it leverages a VLM directly and has no exposure to human preference data, so its scores reflect text-image alignment rather than aesthetic preference. For UnifiedReward, the limitation may stem from its generative interface: when used pointwise, it can only emit integer ratings, providing coarser-grained scores than the continuous outputs of regression-based reward models. While both achieve fairly strong visual reasoning, they still underperform our method (85.3\% average).

\textbf{Ours w/o CrossPairs reproduces the forgetting effect.} Training Qwen3-2B on ImageRewardDB alone yields strong preference accuracy (66.3\%) but degrades visual understanding from 85.4\% (base Qwen3-2B) to 78.4\%---a 7-point drop, with particularly steep losses on Winoground ($-$9.1) and WhatsUp ($-$18.0). This is the same forgetting effect we documented in Section~\ref{sec:continual}.

\textbf{CrossPairs achieves a Pareto improvement.} Our full method matches the preference accuracy of `Ours w/o CrossPairs' (66.3\%) while recovering nearly all of the visual reasoning of the base Qwen3-2B. The breakdown reveals that the CrossPairs actually sharpens counting abilities beyond the base model, while largely preserving the other visual understanding categories.

\textbf{Pointwise evaluation.} All reward models are evaluated in a pointwise fashion: the reward score on a single prompt--image pair is used for ranking, rather than passing both images simultaneously to models that support pairwise scoring. This ensures a fair comparison across architectures and reflects the practical deployment setting, where pointwise scores are easier to leverage for filtering, RL, and inference-time scaling than relative pairwise judgments. UnifiedReward-Think is the one model in Table~\ref{tab:acc_rms_new} not trained for pointwise use; we obtain pointwise ratings from it by prompting, as described above, and include it in the accuracy table only.

\subsection{Further Generators}
\label{appendix:further_generators}

\begin{figure}
\includegraphics[width=\textwidth]{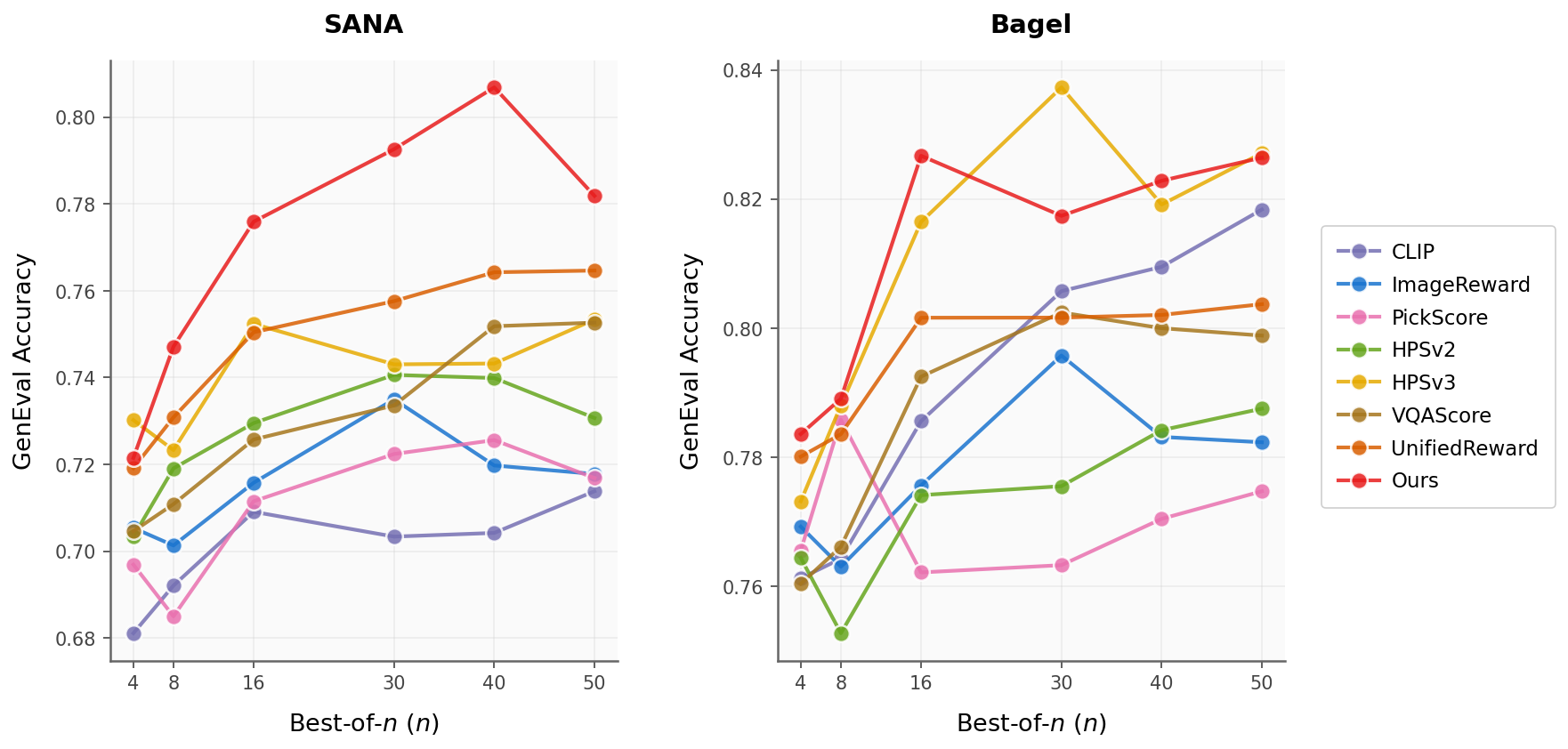}
\caption{Further generators.}
\label{fig:bon_further_generators}
\end{figure}

To assess whether our reward model's advantages generalize across generator architectures, we extend the inference-time best-of-$n$ evaluation from Figure~\ref{fig:geneval_bon} (SD-Turbo and FLUX.1-Schnell) to two additional generators: SANA~\citep{xie2024sana} (specifically, the \texttt{Sana\_1600M\_512px\_diffusers} checkpoint) and Bagel~\citep{deng2025emerging} (specifically, the \texttt{BAGEL-7B-MoT} checkpoint). Together, these four generators span a range of generative modeling paradigms---diffusion (SD-Turbo), rectified flow (FLUX.1-Schnell, SANA), and unified multimodal next-token prediction (Bagel)---as well as a range of architectural backbones: a U-Net (SD-Turbo), an MM-DiT (FLUX.1-Schnell), a linear-attention DiT with deep-compression autoencoder (SANA), and a Mixture-of-Transformer-Experts (Bagel). All evaluations use the same GenEval prompt set as in the main body, with SANA and Bagel called at 8 inference steps to test whether reward model gains hold under the more capable multi-step regime.

Figure~\ref{fig:bon_further_generators} shows GenEval accuracy as a function of $n$ for SANA and Bagel. Several observations:

\textbf{Our reward model dominates on SANA across the curve.} On SANA, our reward model reaches $>$80\% GenEval accuracy at $n=40$, compared to $\sim$76\% for the next-best baseline (UnifiedReward). The gap is consistent throughout the $n$-scaling curve, with our model leading from $n=8$ onward.

\textbf{On Bagel, gains are smaller but consistent.} Bagel starts from a stronger base, leaving less headroom for reward-guided selection. Our reward model still ranks among the top performers across $n$, reaching $\sim$83\% at $n=50$, competitive with HPSv3 and ahead of the remaining baselines.

\textbf{The relative ordering of baselines varies by generator.} While our reward model is consistently the strongest across all four generators (SD-Turbo, FLUX.1-Schnell, SANA, Bagel), the relative ordering among the baselines is inconsistent: a method that ranks among the top on one generator may fall to the middle of the pack on another. 

\subsection{Further evaluations on T2ICompBench}
\label{appendix:t2icompbench}

\begin{figure}[h]
\includegraphics[width=\textwidth]{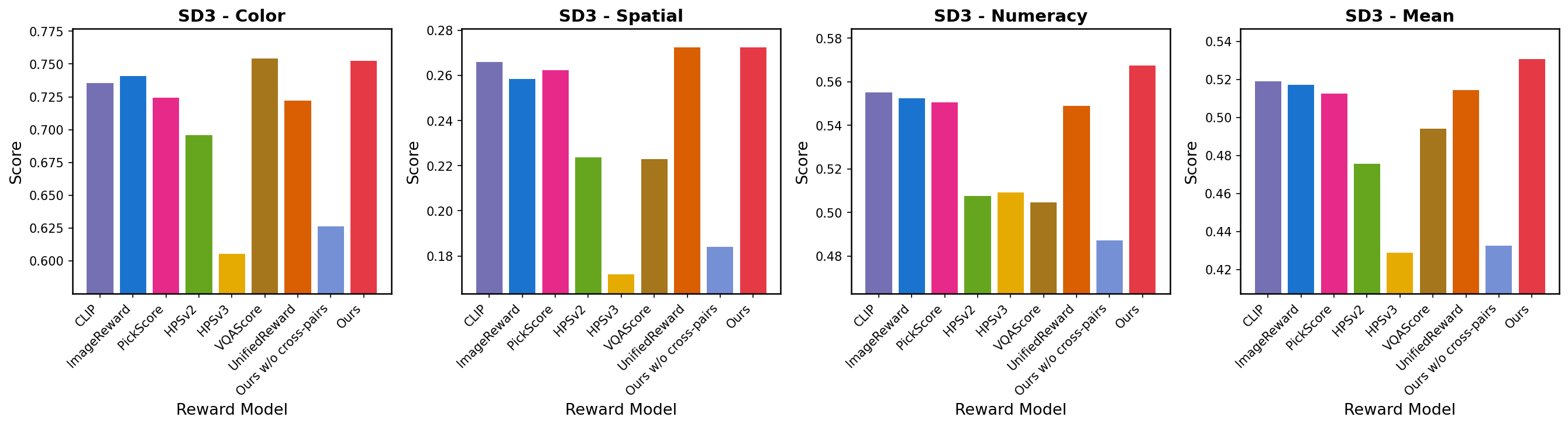}
\caption{Further evaluations on T2ICompBench.}
\label{fig:t2icompbench}
\end{figure}

To test whether the gains on GenEval transfer to a different compositional benchmark, we evaluate the final RL-trained checkpoints on T2I-CompBench~\citep{huang2025t2i}. We sample 300 prompts uniformly at random from each of the color, spatial, and numeracy splits, and evaluate the SD 3.5 Medium model RL-trained with each reward model under DiffusionNFT (the same checkpoints evaluated in Section~\ref{sec:datacentric_results}).
We use the recommended default evaluators from T2I-CompBench: BLIP-VQA for the color split, and UniDet for the spatial and numeracy splits.

Figure~\ref{fig:t2icompbench} reports T2I-CompBench scores across the three splits and their mean. Our reward model achieves the highest mean T2I-CompBench score across the three categories, consistent with our GenEval results in Figure~\ref{fig:combined_geneval}.
The advantage is most pronounced on the spatial and numeracy splits---the categories most directly targeted by our CrossPairs training data (SpatialSense and VSR for spatial; TallyQA for numeracy). On the color split, our reward model performs comparably to the strongest baselines, suggesting that color-attribute binding (targeted by ColorSwap in our mix) is also strengthened, though margins are tighter on this category.

`Ours w/o CrossPairs' underperforms across all splits, again confirming that the data intervention---not the choice of base model or training recipe---drives the improvements.

\subsection{Ablating out CrossPairs}
\label{appendix:ablating_out_crosspairs}

To isolate the contribution of the CrossPairs data intervention from other choices in our pipeline (base model, training hyperparameters, regression head design), we train an ablated model `Ours w/o CrossPairs' on ImageRewardDB preference pairs only, holding all other settings identical to the full method.

\textbf{Reward model accuracy (Table~\ref{tab:acc_rms_new}). }
The ablated model achieves 66.3\% on ImageRewardDB---matching our full method---but drops to 78.4\% average visual understanding (vs.\ 85.3\% for the full method). This reproduces the forgetting pattern documented in Section~\ref{sec:continual}: training on conventional preference pairs alone, even from a strong VLM backbone, erodes visual reasoning. The drop is largest on Winoground ($-$4.8), CV-Relation ($-$19.0 from base), and WhatsUp ($-$18.0 from base).

\textbf{Alignment preferences on generated images.}
The visual understanding evaluations in Table~\ref{tab:acc_rms_new} have the same shared-image, constructed-negative structure of CrossPairs training data, albeit on different (out-of-distribution) datasets. MJ-Bench~\citep{chen2026mj} provides a complementary test: its alignment split consists of pairs of \emph{generated} images under a shared prompt, with human labels for which image better satisfies the prompt. The images carry the artifacts of real generators and the preferred image differs from the dispreferred one in varied and often subtle ways, so neither the image distribution nor the failure modes resemble our rule-based perturbations of real photographs. Table~\ref{tab:mjbench} shows that CrossPairs improve accuracy on every category, by 5.6 points overall and by 8.9 and 7.9 points on Counting and Spatial, the skills most directly targeted by our training sources. The gain on Action, a category none of our perturbations target, suggests the improvement is not limited to the perturbation types seen in training.

\begin{table}[h]
  \caption{Accuracy (\%) on the alignment split of MJ-Bench.}
  \label{tab:mjbench}
  \begin{center}
    \begin{footnotesize}
    \setlength{\tabcolsep}{4pt}
    \begin{tabular}{lrrrrrr}
    \toprule
      & Overall & Object & Attribute & Counting & Spatial & Action \\
    \midrule
    Ours w/o CrossPairs & 62.4 & 61.0 & 62.0 & 66.1 & 67.1 & 61.5 \\
    Ours                & \textbf{68.0} & \textbf{65.7} & \textbf{66.4} & \textbf{75.0} & \textbf{75.0} & \textbf{68.4} \\
    \bottomrule
    \end{tabular}
    \end{footnotesize}
  \end{center}
\end{table}

\begin{figure}[h]
    \centering
    \begin{subfigure}[t]{0.4\columnwidth}
        \centering
        \includegraphics[width=\linewidth]{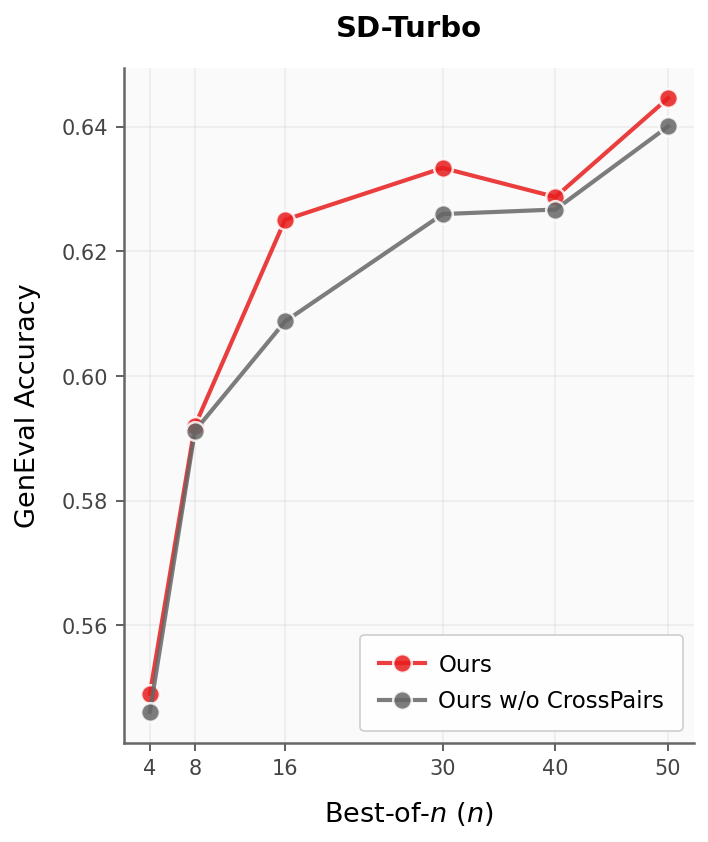}
        \caption{}
    \end{subfigure}
    \begin{subfigure}[t]{0.4\columnwidth}
        \centering
        \includegraphics[width=\linewidth]{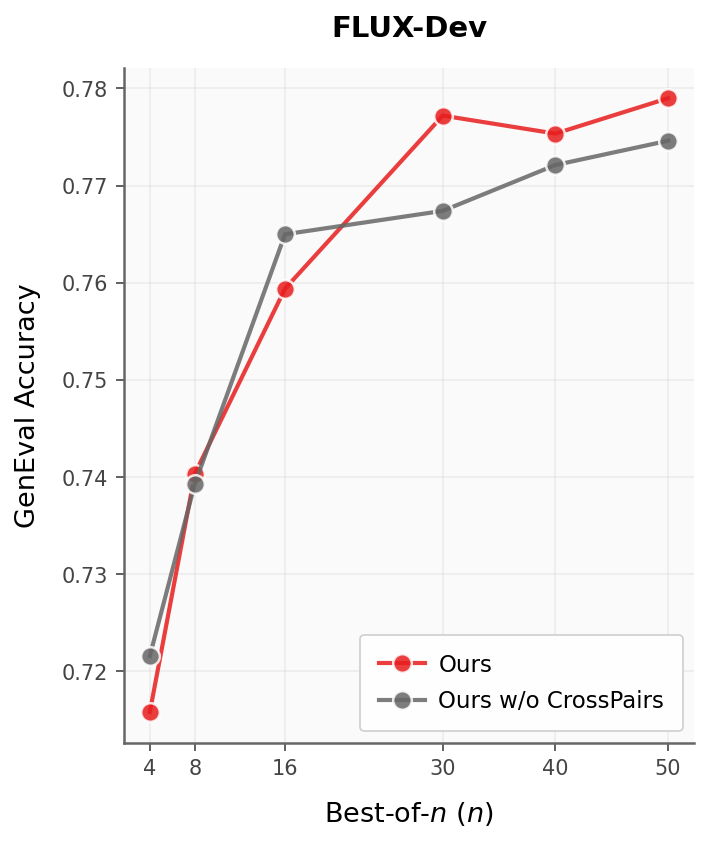}
        \caption{}
    \end{subfigure}
    \begin{subfigure}[t]{0.4\columnwidth}
        \centering
        \includegraphics[width=\linewidth]{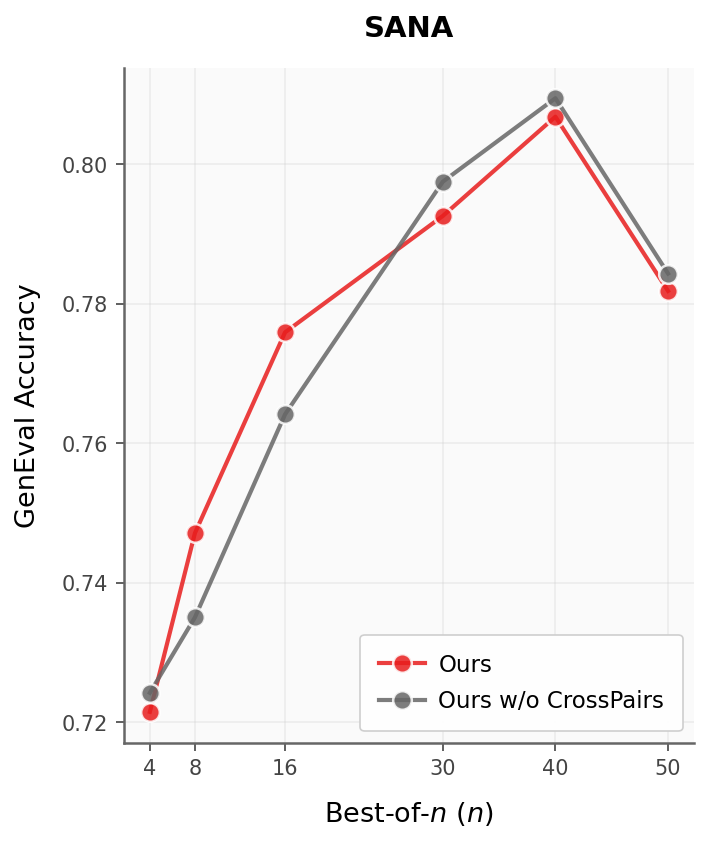}
        \caption{}
    \end{subfigure}
    \begin{subfigure}[t]{0.4\columnwidth}
        \centering
        \includegraphics[width=\linewidth]{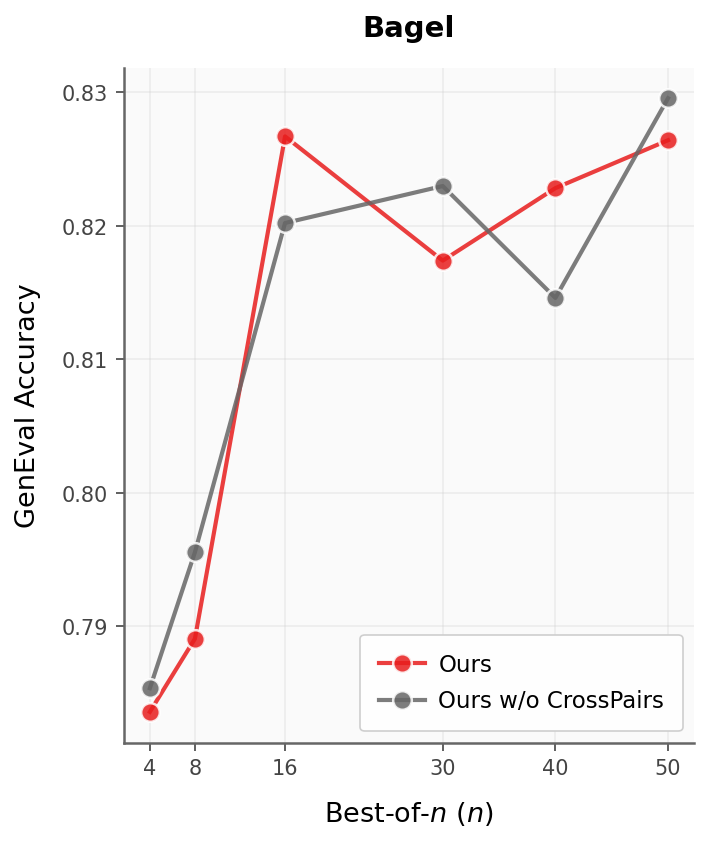}
        \caption{}
    \end{subfigure}
    \caption{Ablating out CrossPairs, evaluated best-of-$n$.}
    \label{fig:ablate_bon}
\end{figure}

\begin{figure}[h]
    \centering
    \begin{subfigure}[t]{0.4\columnwidth}
        \centering
        \includegraphics[width=\linewidth]{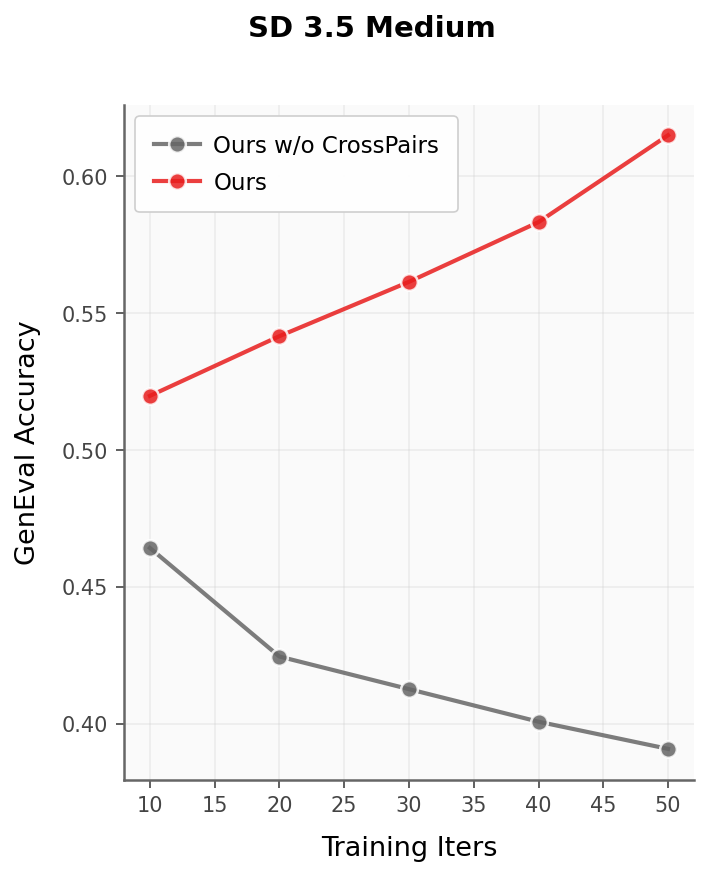}
        \caption{}
    \end{subfigure}
    \begin{subfigure}[t]{0.4\columnwidth}
        \centering
        \includegraphics[width=\linewidth]{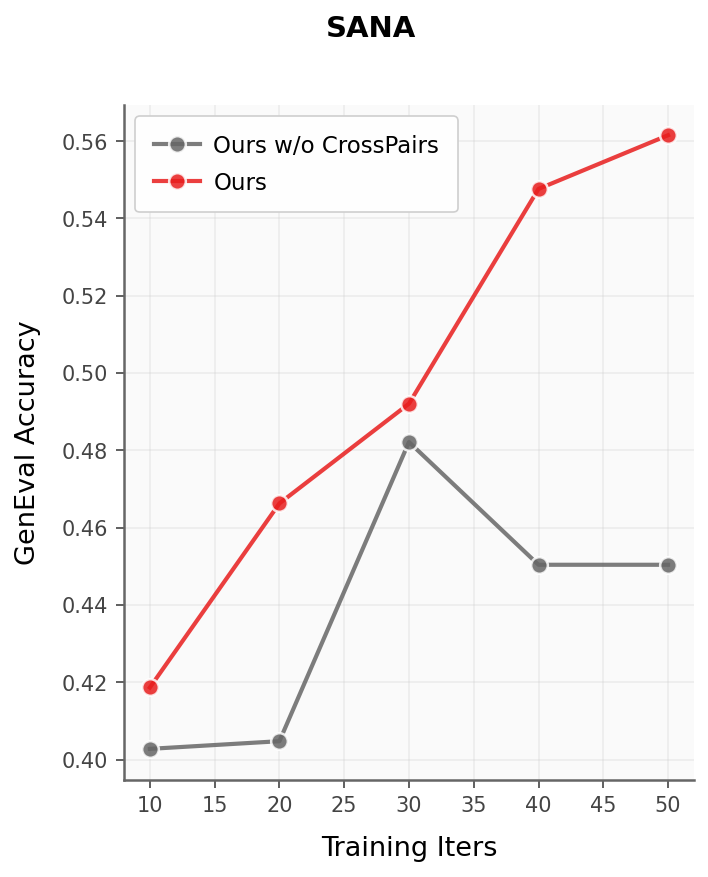}
        \caption{}
    \end{subfigure}
    \caption{Ablating out CrossPairs, evaluated by RL post-training.}
    \label{fig:ablate_rl}
\end{figure}

\textbf{Downstream generation.} When used as a reward signal for inference-time best-of-$n$ on SD-Turbo, FLUX.1-Schnell, SANA, and Bagel (Figure~\ref{fig:ablate_bon}), `Ours w/o CrossPairs' performs marginally worse than the full method, though the gap is not always consistent across generators and values of $n$. The picture changes dramatically when the reward model is used as the objective for RL post-training (Figure~\ref{fig:ablate_rl}). On both SD 3.5 Medium and SANA, RL with `Ours w/o CrossPairs' fails to improve compositional accuracy and on SD 3.5 Medium actually \textit{decreases} GenEval performance over training---from $\sim$46\% at iteration 10 to $\sim$39\% at iteration 50. Meanwhile, RL with our full reward model yields steady improvement, reaching 61.5\% on SD 3.5 Medium and $\sim$56\% on SANA.

\subsection{Preservation of Aesthetic Quality}
\label{appendix:aesthetic}

A natural concern with CrossPairs is whether shifting the training signal toward visual reasoning compromises the reward model's ability to capture aesthetic preferences. We evaluate this in two ways: held-out preference accuracy and aesthetic quality of RL-trained generations. In both cases, we restrict comparison to reward models trained on the same human preference data (the ImageReward reward model and our ablated baseline `Ours w/o CrossPairs') to make a fair comparison.

\textbf{Held-out preference accuracy (Table~\ref{tab:held_out_pref_pair_acc}).} We compare ImageReward, `Ours w/o CrossPairs', and our full method on ImageRewardDB (in-distribution) and the 500-pair `unique test' split of Pick-a-Pic~\citep{kirstain2023pick} (out-of-distribution---these models have not been trained on Pick-a-Pic data). On both splits, the CrossPairs intervention matches or exceeds the preference accuracy of the ablated baseline: 66.3\% vs.\ 66.3\% on ImageRewardDB (tied) and 63.6\% vs.\ 63.2\% on Pick-a-Pic. Both variants outperform ImageReward (65.1\% / 60.9\%). The fact that adding CrossPairs supervision does \textit{not} reduce held-out preference accuracy---and slightly improves OOD generalization on Pick-a-Pic---indicates that visual reasoning supervision is complementary to, rather than competitive with, aesthetic preference learning.

\begin{table}[h]
  \centering
  \caption{Held out preference accuracy.}
  \label{tab:datasets}
  \begin{tabular}{lcc}
    \toprule
    & ImageRewardDB (ID) & Pick-a-Pic (OOD) \\
    \midrule
    ImageReward & 65.1 & 60.9 \\
    Ours w/o CrossPairs & 66.3 & 63.2 \\
    Ours & 66.3 & 63.6 \\
    \bottomrule
  \end{tabular}
  \label{tab:held_out_pref_pair_acc}
\end{table}

\textbf{Aesthetic score on RL-trained generations (Table~\ref{tab:aesthetic_scores}).} 
As an alternative test, we apply the LAION aesthetic predictor to images generated by SD 3.5 Medium and SANA models RL-trained against each of the three reward models, evaluated on 512 held-out prompts from the Pick-a-Pic test set. The differences are marginal, all within $\sim$0.1 of each other. For context, LAION uses this same predictor to bucket the LAION-5B English subset into aesthetic tiers, with the 6.5+ tier (625K image-text pairs) being a strict subset of the 4.5+ tier (1.2B pairs) commonly used for fine-tuning generative models. All three reward models thus produce generations in the strong-aesthetics regime, with differences smaller than the gap between adjacent LAION tiers.

\textbf{Qualitative confirmation.} The qualitative examples in Section~\ref{appendix:further_examples} further support that aesthetic quality is preserved: generations from our reward model are visually crisp and compositionally polished, while the prompts from PickScore (first three columns) show no obvious aesthetic regression compared to ImageReward and HPSv3 baselines.

Together, these results indicate that CrossPairs improves visual reasoning without sacrificing the aesthetic preference signal.

\begin{table}[h]
  \centering
  \caption{Aesthetic scores on generations from reinforcement post-trained generators.}
  \label{tab:base_models}
  \begin{tabular}{lcc}
    \toprule
    & SD 3.5 Medium & SANA \\
    \midrule
    ImageReward & 6.46 & 6.48 \\
    Ours w/o CrossPairs & 6.42 & 6.50 \\
    Ours & 6.42 & 6.41 \\
    \bottomrule
  \end{tabular}
  \label{tab:aesthetic_scores}
\end{table}

\subsection{Further Qualitative Examples}
\label{appendix:further_examples}

\begin{figure}
\includegraphics[width=\textwidth]{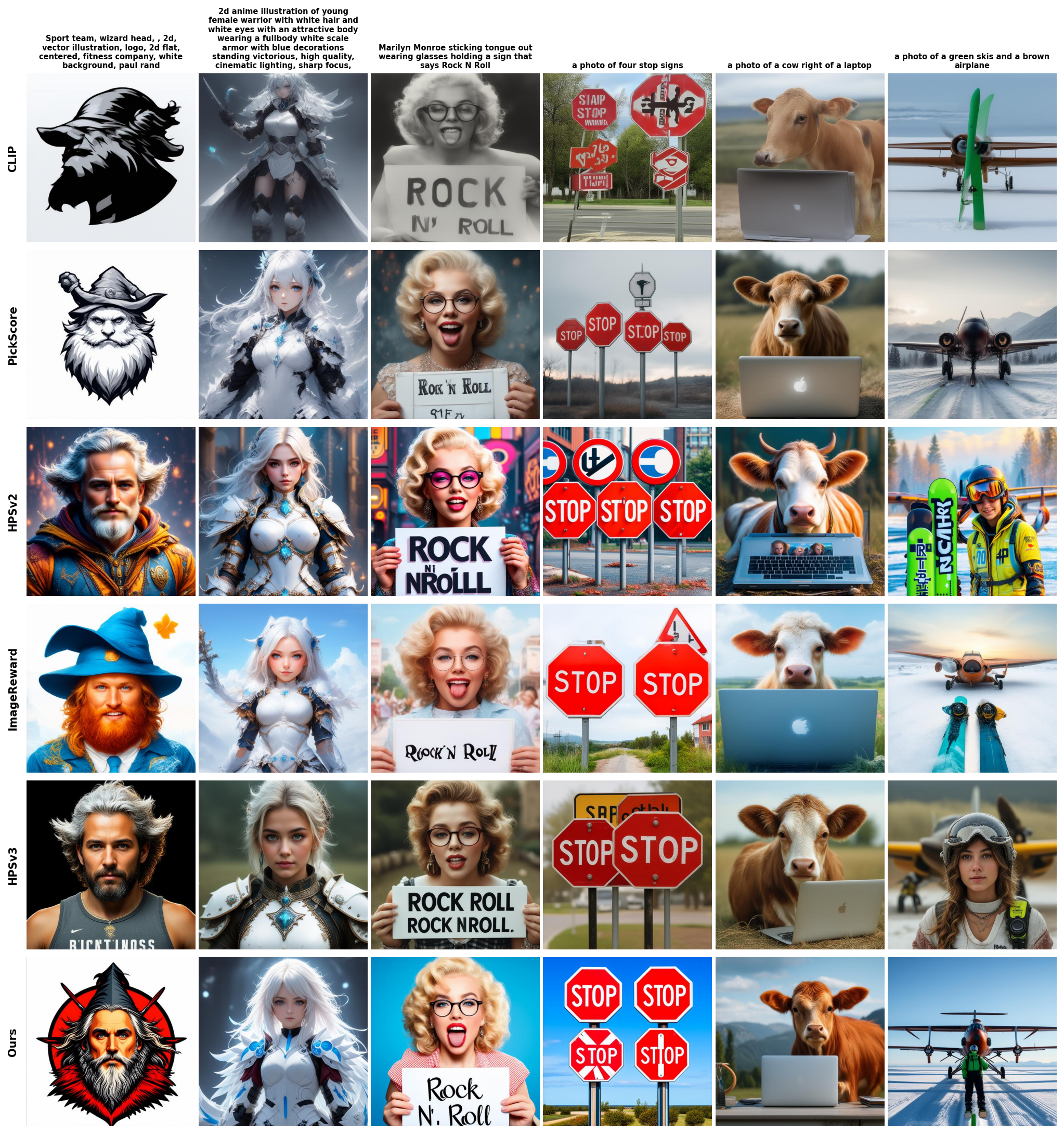}
\caption{Example generations. The first three columns are prompts from PickScore; the last three columns are prompts from GenEval.}
\label{fig:examples_all_rms}
\end{figure}

\begin{figure}
\includegraphics[width=\textwidth]{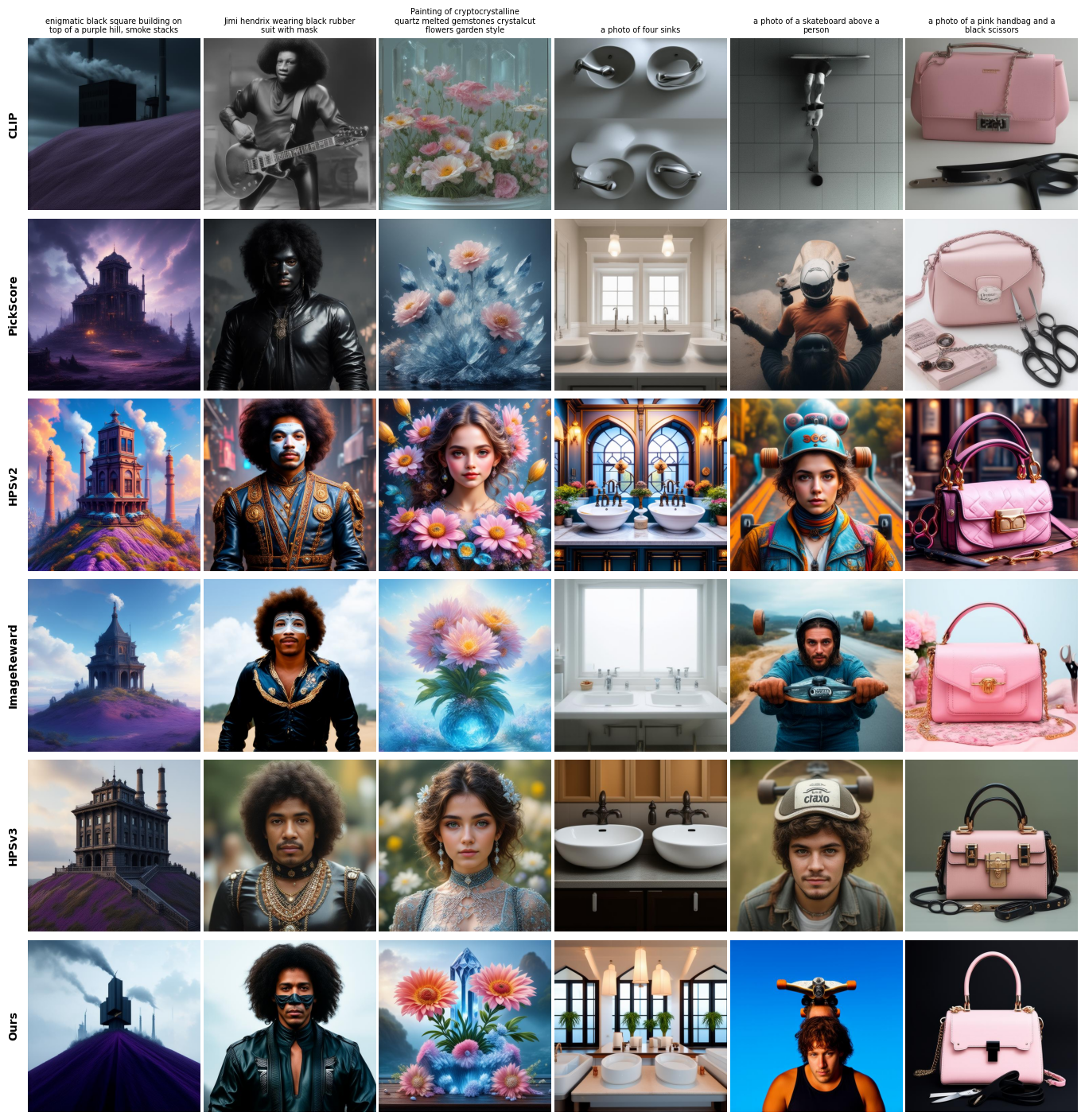}
\caption{Example generations. The first three columns are prompts from PickScore; the last three columns are prompts from GenEval.}
\label{fig:examples_all_rms_moree}
\end{figure}

See Figures \ref{fig:examples_all_rms} and \ref{fig:examples_all_rms_moree}.

\section{CrossPairs vs.\ a Next-Token-Prediction Loss}
\label{app:vqa_aux}

CrossPairs inject visual reasoning supervision into preference learning. A
natural question is whether the \emph{pairing structure} matters, or whether any
form of visual reasoning supervision drawn from the same sources would do. Since
our backbone is a VLM, the most obvious alternative is to keep the
Bradley--Terry objective on ImageRewardDB and add the visual reasoning data as
an auxiliary next-token-prediction (NTP) loss on the language modeling head,
i.e.\ ordinary VQA fine-tuning.

\paragraph{Setup.} We hold the backbone (Qwen3-2B), trainable layers, readout
head, hyperparameters, and the data mixture of Table~\ref{tab:data_mix} fixed,
and vary only how the visual reasoning sources are supervised. For the
\emph{VQA auxiliary loss} variant, each source is used in its original,
unperturbed VQA form with the usual language modeling loss: TallyQA count
questions completed by the numerical answer; VSR and SpatialSense statements
completed by \texttt{True}/\texttt{False}; ColorSwap as an A/B caption choice;
and CLEVR as the correct caption describing the scene. ImageRewardDB pairs are
supervised through the regression head with the Bradley--Terry loss as in our
main results. For \emph{Ours}, the same sources are converted into CrossPairs
(Section~\ref{sec:datacentric}) and supervised through the same Bradley--Terry
loss and regression head as the preference pairs.

\begin{table*}[h]
  \caption{CrossPairs vs.\ supervising the same visual reasoning sources with an
  auxiliary VQA next-token-prediction loss. Accuracy on ImageRewardDB and the
  visual understanding evaluations; random performance is 50\%.}
  \label{tab:vqa_aux_acc}
  \begin{center}
    \begin{footnotesize}
    \setlength{\tabcolsep}{1.5pt}
    \begin{tabular}{lccrrrrrrrr}
    \toprule
      &
      \multirow{2}{*}{\centering ImageRDB} &
      \multirow{2}{*}{\centering Visual} &
      \multicolumn{8}{c}{Visual Understanding Breakdown} \\
      \cmidrule(lr){4-11}
      & & &
      Wground & SC-Add & SC-Rep & SC-Swap & CV-Rel & WhatsUp & CV-Count & CountBch \\
    \midrule
    VQA auxiliary loss
      & 65.2 & 82.2 & 66.6 & 97.6 & 90.7 & 89.9 & 77.1 & 80.8 & 67.7 & 87.2 \\
    Ours
      & 66.3 & 85.3 & 71.6 & 97.5 & 92.8 & 91.9 & 74.5 & 92.2 & 71.7 & 90.2 \\
    \bottomrule
    \end{tabular}
    \end{footnotesize}
  \end{center}
\end{table*}

\begin{table}[h]
  \caption{Best-of-$n$ GenEval accuracy (\%) for the two supervision formats.
  We generate 50 images per prompt; each entry is averaged over 200 random
  $n$-subsets of those 50 generations.}
  \label{tab:vqa_aux_bon}
  \begin{center}
    \begin{footnotesize}
    \setlength{\tabcolsep}{4pt}
    \begin{tabular}{llrrrrrr}
    \toprule
      Generator & Method & $n{=}4$ & $n{=}8$ & $n{=}16$ & $n{=}30$ & $n{=}40$ & $n{=}50$ \\
    \midrule
    \multirow{2}{*}{SD-Turbo}
      & VQA auxiliary loss & 55.4 & 58.5 & 60.8 & 62.4 & 63.0 & 63.4 \\
      & Ours               & 55.4 & 58.5 & 61.0 & 63.0 & 63.8 & 64.5 \\
    \midrule
    \multirow{2}{*}{FLUX.1-Schnell}
      & VQA auxiliary loss & 71.9 & 73.8 & 75.0 & 75.9 & 76.4 & 76.9 \\
      & Ours               & 71.8 & 73.7 & 75.2 & 76.6 & 77.3 & 78.1 \\
    \bottomrule
    \end{tabular}
    \end{footnotesize}
  \end{center}
\end{table}

\paragraph{Results.} CrossPairs are consistently better on the joint objective
(Table~\ref{tab:vqa_aux_acc}): higher ImageRewardDB accuracy (66.3\% vs.\
65.2\%) and higher average visual understanding (85.3\% vs.\ 82.2\%), with the
largest gaps on WhatsUp ($+11.4$) and Winoground ($+5.0$); CV-Relation is the one
split on which the auxiliary loss is marginally ahead. Downstream
(Table~\ref{tab:vqa_aux_bon}), the two are indistinguishable up to $n{=}16$ on
both generators, after which CrossPairs pull ahead and the gap widens with $n$.

We note finally that CrossPairs are, in our view, the most natural way to
leverage VQA-style supervision for reward modeling. A CrossPair is already a
valid Bradley--Terry pair (the two candidates differ along the text axis rather
than the image axis), so it is consumed by the same loss and readout head as the
human preference data, with no auxiliary objective, loss weighting, or language
modeling head to maintain. The same construction therefore applies to backbones
that cannot generate text at all, such as CLIP-style dual encoders, for which
the NTP alternative is not available.
\section{Error analysis}
\label{app:bon_variance}

The reward model accuracies in Tables~\ref{tab:acc_rms_new} and \ref{tab:vqa_aux_acc} are deterministic: each is a single pass of a deterministic reward model over a fixed set of inputs, so there is no run-to-run variance. In contrast, the best-of-$n$ results admit an error analysis based on random redraws of the $n$ candidates at each point of the curve. Generating fresh candidates for every $n$ and every reward model is beyond our compute budget, so we use the following resampling protocol. For each prompt we generate 50 images once; for each $n$ we draw 200 random $n$-subsets of those 50, apply best-of-$n$ selection within each subset, and report the mean and standard deviation across draws. The draws are not independent samples from the generator, so this underestimates the true variance, but it captures the selection noise. At $n{=}50$ the subset is the full set and the standard deviation is zero by construction.\footnote{
Except in the cases of argmax tie-breaking, where multiple samples attain the same reward but score differently on downstream evaluation.}

\begin{table}[h]
  \caption{Best-of-$n$ GenEval accuracy (\%, mean $\pm$ std over 200 resampled
  $n$-subsets) with SD-Turbo as the generator.}
  \label{tab:bon_sdturbo}
  \begin{center}
    \begin{footnotesize}
    \setlength{\tabcolsep}{3pt}
    \begin{tabular}{lrrrrrr}
    \toprule
      Method & $n{=}4$ & $n{=}8$ & $n{=}16$ & $n{=}30$ & $n{=}40$ & $n{=}50$ \\
    \midrule
    CLIP          & 54.0 $\pm$ 1.3 & 56.1 $\pm$ 1.2 & 57.4 $\pm$ 1.2 & 57.8 $\pm$ 1.0 & 57.8 $\pm$ 0.8 & 57.3 $\pm$ 0.2 \\
    ImageReward   & 54.9 $\pm$ 1.3 & 57.4 $\pm$ 1.2 & 59.5 $\pm$ 1.2 & 60.6 $\pm$ 0.9 & 60.9 $\pm$ 0.6 & 61.2 $\pm$ 0.0 \\
    PickScore     & 53.0 $\pm$ 1.2 & 55.0 $\pm$ 1.2 & 56.7 $\pm$ 1.2 & 58.3 $\pm$ 0.9 & 59.1 $\pm$ 0.7 & 60.0 $\pm$ 0.0 \\
    HPSv2         & 52.9 $\pm$ 1.2 & 54.6 $\pm$ 1.1 & 56.0 $\pm$ 1.0 & 56.8 $\pm$ 0.9 & 57.4 $\pm$ 0.8 & 57.9 $\pm$ 0.0 \\
    HPSv3         & 53.7 $\pm$ 1.2 & 56.2 $\pm$ 1.2 & 58.4 $\pm$ 1.1 & 60.6 $\pm$ 0.9 & 61.5 $\pm$ 0.7 & 62.4 $\pm$ 0.0 \\
    VQAScore      & 53.8 $\pm$ 1.4 & 56.3 $\pm$ 1.3 & 58.3 $\pm$ 1.2 & 59.9 $\pm$ 0.9 & 60.6 $\pm$ 0.8 & 61.1 $\pm$ 0.0 \\
    UnifiedReward & 54.8 $\pm$ 1.4 & 57.3 $\pm$ 1.3 & 59.0 $\pm$ 1.2 & 60.5 $\pm$ 1.1 & 61.2 $\pm$ 1.1 & 62.0 $\pm$ 0.0 \\
    Ours          & \textbf{55.4} $\pm$ 1.3 & \textbf{58.5} $\pm$ 1.2 & \textbf{61.0} $\pm$ 1.2 & \textbf{63.0} $\pm$ 1.0 & \textbf{63.8} $\pm$ 0.8 & \textbf{64.5} $\pm$ 0.0 \\
    \bottomrule
    \end{tabular}
    \end{footnotesize}
  \end{center}
\end{table}

\begin{table}[h]
  \caption{Best-of-$n$ GenEval accuracy (\%, mean $\pm$ std over 200 resampled
  $n$-subsets) with FLUX.1-Schnell as the generator.}
  \label{tab:bon_flux}
  \begin{center}
    \begin{footnotesize}
    \setlength{\tabcolsep}{3pt}
    \begin{tabular}{lrrrrrr}
    \toprule
      Method & $n{=}4$ & $n{=}8$ & $n{=}16$ & $n{=}30$ & $n{=}40$ & $n{=}50$ \\
    \midrule
    CLIP          & 70.8 $\pm$ 1.1 & 72.2 $\pm$ 1.2 & 73.4 $\pm$ 1.1 & 74.4 $\pm$ 0.9 & 74.9 $\pm$ 0.6 & 75.7 $\pm$ 0.0 \\
    ImageReward   & 71.1 $\pm$ 1.1 & 72.7 $\pm$ 1.1 & 73.8 $\pm$ 1.0 & 74.7 $\pm$ 0.9 & 74.9 $\pm$ 0.7 & 74.9 $\pm$ 0.0 \\
    PickScore     & 70.1 $\pm$ 1.1 & 71.0 $\pm$ 1.1 & 71.5 $\pm$ 1.0 & 71.7 $\pm$ 0.8 & 71.7 $\pm$ 0.6 & 71.6 $\pm$ 0.0 \\
    HPSv2         & 70.0 $\pm$ 1.1 & 71.1 $\pm$ 1.1 & 71.7 $\pm$ 1.0 & 71.9 $\pm$ 0.8 & 71.9 $\pm$ 0.6 & 71.8 $\pm$ 0.0 \\
    HPSv3         & 71.2 $\pm$ 1.1 & 72.9 $\pm$ 1.1 & 74.2 $\pm$ 1.0 & 75.1 $\pm$ 0.8 & 75.4 $\pm$ 0.6 & 75.5 $\pm$ 0.0 \\
    VQAScore      & 70.2 $\pm$ 1.2 & 71.7 $\pm$ 1.2 & 73.1 $\pm$ 1.1 & 74.3 $\pm$ 1.0 & 74.8 $\pm$ 0.8 & 75.3 $\pm$ 0.6 \\
    UnifiedReward & 70.9 $\pm$ 1.2 & 72.4 $\pm$ 1.2 & 73.4 $\pm$ 1.2 & 74.1 $\pm$ 1.1 & 74.3 $\pm$ 1.1 & 74.5 $\pm$ 1.0 \\
    Ours          & \textbf{71.8} $\pm$ 1.1 & \textbf{73.7} $\pm$ 1.1 & \textbf{75.2} $\pm$ 1.0 & \textbf{76.6} $\pm$ 0.9 & \textbf{77.3} $\pm$ 0.7 & \textbf{78.1} $\pm$ 0.0 \\
    \bottomrule
    \end{tabular}
    \end{footnotesize}
  \end{center}
\end{table}

Our reward model leads on both generators at every $n$. At small $n$ the margin over the strongest baseline (UnifiedReward on SD-Turbo, HPSv3 on
FLUX.1-Schnell) is within sampling noise; it exceeds two standard deviations from $n{=}16$ onward on SD-Turbo and from $n{=}30$ onward on FLUX.1-Schnell, and widens further as $n$ grows.

\section{Generalization to Another VLM Backbone}
\label{app:smolvlm}

All experiments in the main body use Qwen3-2B as the reward model backbone. To demonstrate the generality of the findings, we repeat the core comparison with SmolVLM2-2.2B~\citep{marafioti2025smolvlm}. We chose it as a widely used open-source VLM of similar size that differs from Qwen3-2B in most design choices: language backbone, multimodal training recipe, tiled fixed-resolution vs dynamic-resolution visual inputs, with / without DeepStack injection of visual features, etc. We use the same data mix, training protocol, and hyperparameters as for Qwen3-2B, and train the same two variants as in Appendix~\ref{appendix:ablating_out_crosspairs}: preference pairs only (\emph{w/o CrossPairs}) and the full mix (\emph{Ours}). We did not run RL post-training with these reward models due to compute constraints.

\begin{table*}[t]
  \caption{SmolVLM2-2.2B as the reward model backbone. Accuracy on ImageRewardDB
  and the visual understanding evaluations; random performance is 50\%.}
  \label{tab:smolvlm_acc}
  \begin{center}
    \begin{footnotesize}
    \setlength{\tabcolsep}{1.5pt}
    \begin{tabular}{lccrrrrrrrr}
    \toprule
      &
      \multirow{2}{*}{\centering ImageRDB} &
      \multirow{2}{*}{\centering Visual} &
      \multicolumn{8}{c}{Visual Understanding Breakdown} \\
      \cmidrule(lr){4-11}
      & & &
      Wground & SC-Add & SC-Rep & SC-Swap & CV-Rel & WhatsUp & CV-Count & CountBch \\
    \midrule
    SmolVLM2-2.2B (base)
      & 55.7 & 76.1 & 69.7 & 93.1 & 85.5 & 83.3 & 69.0 & 76.4 & 62.6 & 69.7 \\
    w/o CrossPairs
      & 58.4 & 69.4 & 62.8 & 89.4 & 86.8 & 71.4 & 55.5 & 59.8 & 59.3 & 70.5 \\
    Ours
      & 58.8 & 73.6 & 65.1 & 91.3 & 90.1 & 76.5 & 62.9 & 66.7 & 65.5 & 70.3 \\
    \bottomrule
    \end{tabular}
    \end{footnotesize}
  \end{center}
\end{table*}

\begin{table}[t!]
  \caption{Best-of-$n$ GenEval accuracy (\%, mean $\pm$ std over 200 resampled
  $n$-subsets; see Appendix~\ref{app:bon_variance}) with SmolVLM2-2.2B-based
  reward models.}
  \label{tab:smolvlm_bon}
  \begin{center}
    \begin{footnotesize}
    \setlength{\tabcolsep}{3pt}
    \begin{tabular}{llrrrrrr}
    \toprule
      Generator & Method & $n{=}4$ & $n{=}8$ & $n{=}16$ & $n{=}30$ & $n{=}40$ & $n{=}50$ \\
    \midrule
    \multirow{2}{*}{SD-Turbo}
      & w/o CrossPairs & 53.9 $\pm$ 1.3 & 55.7 $\pm$ 1.2 & 57.1 $\pm$ 1.1 & 58.2 $\pm$ 0.9 & 58.3 $\pm$ 0.7 & 57.9 $\pm$ 0.0 \\
      & Ours           & \textbf{54.2} $\pm$ 1.3 & \textbf{56.1} $\pm$ 1.0 & \textbf{57.5} $\pm$ 1.1 & \textbf{58.8} $\pm$ 1.0 & \textbf{59.2} $\pm$ 0.7 & \textbf{59.3} $\pm$ 0.0 \\
    \midrule
    \multirow{2}{*}{FLUX.1-Schnell}
      & w/o CrossPairs & 69.7 $\pm$ 1.1 & 70.8 $\pm$ 1.1 & 71.4 $\pm$ 1.1 & 71.6 $\pm$ 0.9 & 71.8 $\pm$ 0.7 & 72.1 $\pm$ 0.0 \\
      & Ours           & \textbf{70.1} $\pm$ 1.1 & \textbf{71.5} $\pm$ 1.1 & \textbf{72.5} $\pm$ 1.0 & \textbf{73.4} $\pm$ 0.9 & \textbf{73.6} $\pm$ 0.7 & \textbf{73.7} $\pm$ 0.0 \\
    \bottomrule
    \end{tabular}
    \end{footnotesize}
  \end{center}
\end{table}

\paragraph{Results.} The pattern from Qwen3-2B reproduces on SmolVLM2 (Table~\ref{tab:smolvlm_acc}). Preference fine-tuning alone erodes visual understanding from 76.1\% to 69.4\%, with the steepest losses on WhatsUp (-16.6), CV-Relation (-13.5), and SugarCrepe-Swap (-11.9), the same categories that degrade most for Qwen3-2B. Adding CrossPairs recovers much of this loss (73.6\%) with no cost to preference accuracy (58.8\% vs.\ 58.4\%), again with the largest gains on the spatial and counting splits. In best-of-$n$ selection (Table~\ref{tab:smolvlm_bon}), the CrossPairs-trained reward model is ahead at every $n$ on both generators, with the gap widening as $n$ grows, mirroring the behavior of our Qwen3-2B model. We take this as evidence that the forgetting effect and the data-centric remedy are properties of preference fine-tuning rather than of a particular backbone.
\section{Limitations}
\label{appendix:limitations}

Two limitations suggest natural avenues for future work. First, we do not characterize how the CrossPairs mix should scale with preference dataset size or base model capacity. Our experiments fix both and the data mix in Table~\ref{tab:data_mix} reflect natural source sizes rather than a recipe of proportions. Whether the optimal CrossPairs-to-preference ratio remains stable when scaling up remains an open question. Second, we evaluate only the discriminative regression-head paradigm. Whether CrossPairs supervision transfers to generative reward models (which produce verbalized ratings or preference statements as tokens) or pairwise reward models (which condition on both candidates jointly) remains untested. We hypothesize the intervention is largely architecture-agnostic, since the underlying signal — image-anchored prompt discrimination — does not depend on the readout mechanism, but verifying this would broaden the applicability of our findings.

\section{Societal Impacts}
\label{appendix:societal}

Our work improves the prompt adherence of text-to-image generative models, which carries dual-use implications common to advances in image generation. On the positive side, more reliable text-image alignment reduces the trial-and-error burden on users, broadens accessibility for non-expert users who cannot iteratively refine prompts, and supports legitimate applications in education, design, and accessibility. On the negative side, stronger prompt adherence could lower the barrier to producing targeted synthetic media, including misinformation, non-consensual imagery, and content designed to evade safety filters that rely on the imprecision of current generators. 

\FloatBarrier


\end{document}